\documentclass[10pt,twocolumn,letterpaper]{article}

\usepackage{cvpr}              

\graphicspath{{figs/}}
\usepackage{graphicx}
\usepackage{booktabs}
\usepackage{wrapfig}
\usepackage{multirow}
\usepackage{colortbl}
\usepackage{appendix}
\usepackage{placeins}
\usepackage{float}
\usepackage{balance}

\definecolor{cvprblue}{rgb}{0.21,0.49,0.74}
\usepackage[pagebackref,breaklinks,colorlinks,allcolors=cvprblue]{hyperref}

\def\paperID{xxx} 
\def\confName{CVPR}
\def\confYear{xxx}

\title{MINIMA: Modality Invariant Image Matching}

\author{Jiangwei Ren$^{1}$, Xingyu Jiang$^{1\dag}$, Zizhuo Li$^{2}$,  Dingkang Liang$^{1}$, Xin Zhou$^{1}$, Xiang Bai$^{1}$  \\
        $^{1}$ Huazhong University of Science and Technology,
        ~~~$^{2}$ Wuhan University\\
        {\tt \{jwren, jiangxy998, dkliang, xbai\}@hust.edu.cn}\\
}
\begin{document}
\maketitle
\let\thefootnote\relax\footnotetext{$\dag$ Corresponding author.}

\begin{abstract}
Image matching for both cross-view and cross-modality plays a critical role in multimodal perception. In practice, the modality gap caused by different imaging systems/styles poses great challenges to the matching task. Existing works try to extract invariant features for specific modalities and train on limited datasets, showing poor generalization. In this paper, we present MINIMA, a unified image matching framework for multiple cross-modal cases. Without pursuing fancy modules, our MINIMA aims to enhance universal performance from the perspective of data scaling up. For such purpose, we propose a simple yet effective data engine that can freely produce a large dataset containing multiple modalities, rich scenarios, and accurate matching labels. Specifically, we scale up the modalities from cheap but rich RGB-only matching data, by means of generative models. Under this setting, the matching labels and rich diversity of the RGB dataset are well inherited by the generated multimodal data. Benefiting from this, we construct MD-syn, a new comprehensive dataset that fills the data gap for general multimodal image matching. With MD-syn, we can directly train any advanced matching pipeline on randomly selected modality pairs to obtain cross-modal ability.
Extensive experiments on in-domain and zero-shot matching tasks, including $19$ cross-modal cases, demonstrate that our MINIMA can significantly outperform the baselines and even surpass modality-specific methods. The dataset and code are available at \url{https://github.com/LSXI7/MINIMA}.

\end{abstract}

\begin{figure}
  \centering
  \includegraphics[width=0.48\linewidth]{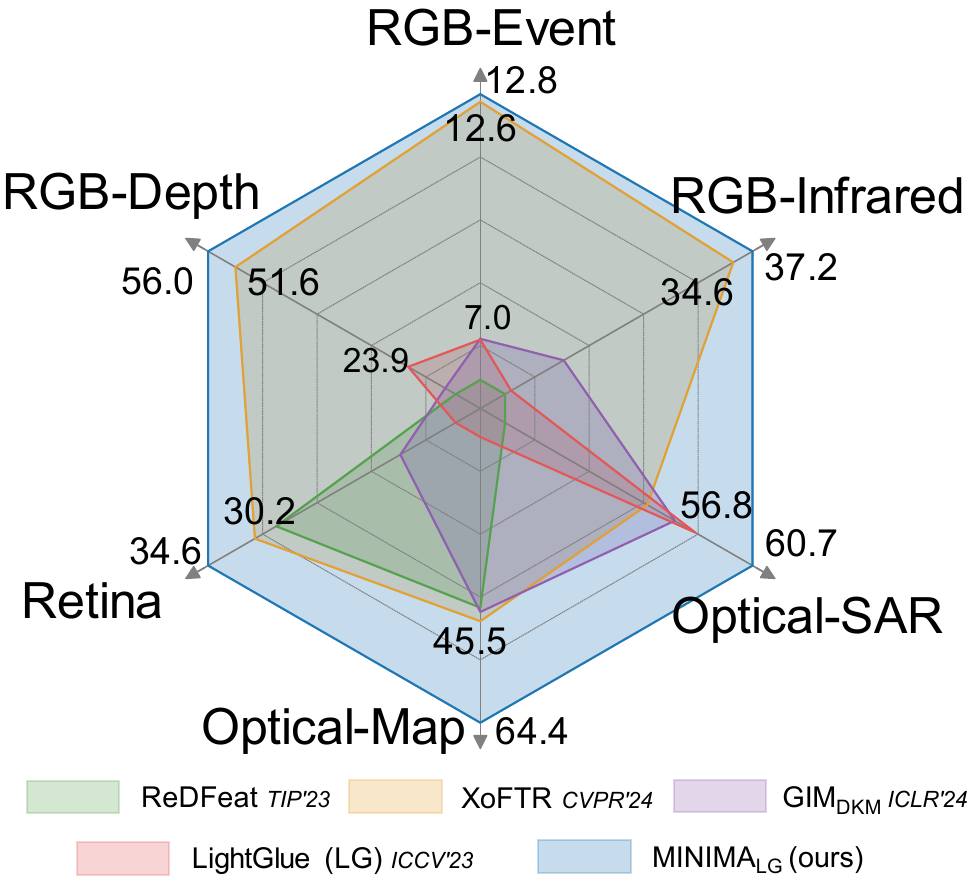}
  \includegraphics[width=0.50\linewidth]{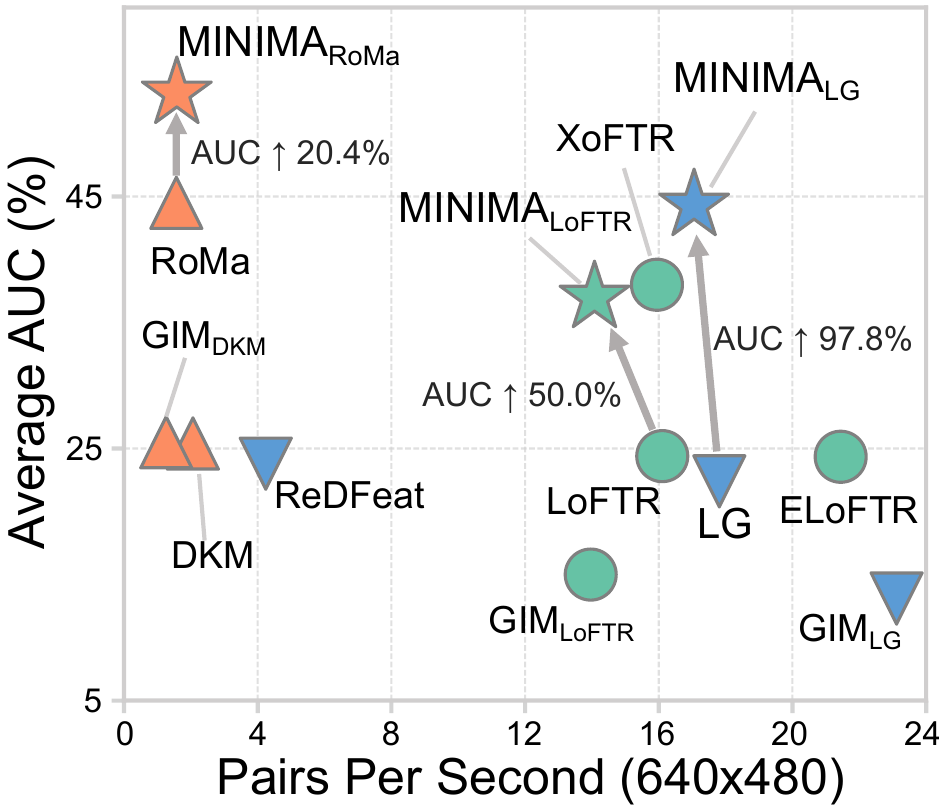}
  \caption{\textbf{Overall Image Matching Accuracy and Efficiency on Six Datasets of Real Cross-modal Image Pairs.} AUC of the pose error (@$10^\circ$) or reprojection error (@$10$px) is used for accuracy evaluation, while Pairs Per Second is used for efficiency test. \textit{Left}: AUCs on each dataset of representative methods are reported. \textit{Right}: average performance is summarized, wherein different colors indicate matching pipelines of sparse, semi-dense, and dense matching, while our MINIMA is marked as $\bigstar$. Using only synthetic multimodal data created by our data engine, MINIMA can generalize to real cross-modal scenes with large improvements. }\label{fig1:overall-performance}
\end{figure}

\section{Introduction}
\label{sec:intro}
Image matching refers to establishing pixel-wise correspondences from two-view images, which serves as a prerequisite for a wide range of visual applications~\cite{ma2021image}. Recently, matching two images of different imaging systems/styles plays a vital role in multimodal perceptions~\cite{jiang2021review}, including image fusion and enhancement~\cite{xu2023murf,zhang2021image}, visual localization/navigation~\cite{zhou2021vmloc,aditya2024thermal}, target detection/recognition/tracking~\cite{gehrig2024low,wu2024single,zhu2023visual,zhang2023self,zhang2023diverse,yang2023towards}, \emph{etc}. They benefit from gathering the advantages of different modalities by aligning them, thereby yielding more comprehensive representations. 
However, the cross-view and cross-modality nature makes the matching task more challenging, particularly using a single model for different modalities such as \emph{RGB-Infrared (IR), RGB-Depth, and RGB-Event}.

\def\ss{10}
\begin{figure*}[h!]
    \centering \vspace{-3pt}
    \resizebox{\textwidth}{!}{
   \includegraphics[width=0.95\linewidth]{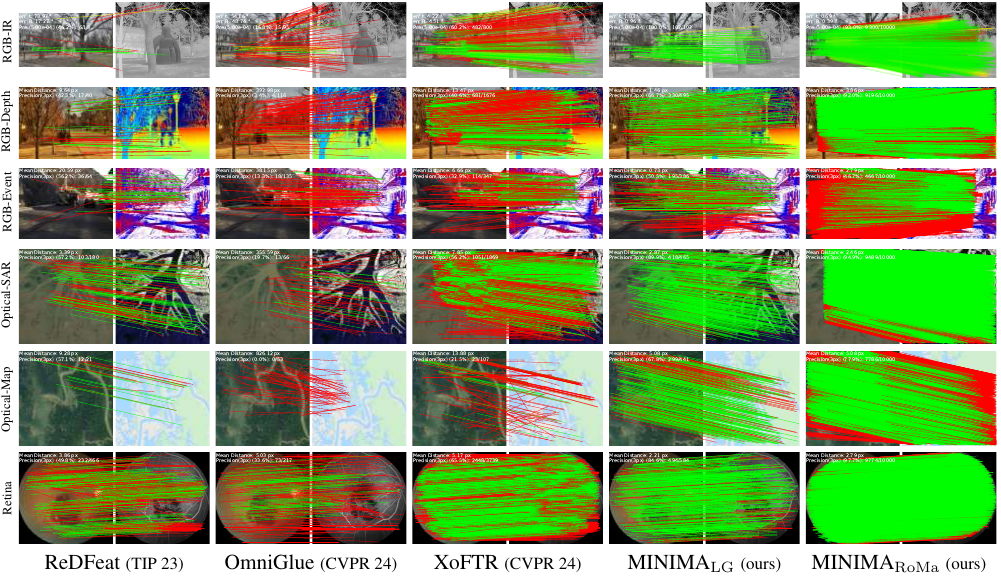}
    }
    \caption{\textbf{Qualitative Results on Real Cross-modal Image Pairs.} Our methods MINIMA$_\mathrm{LG}$ (sparse) and MINIMA$_\mathrm{RoMa}$ (dense) are compared with the sparse matching pipeline ReDFeat~\cite{deng2022redfeat} and OmniGlue~\cite{jiang2024omniglue}, and semi-dense matcher XoFTR~\cite{tuzcuouglu2024xoftr}. ReDFeat and XoFTR are cross-modal methods, and OmniGlue is known for its generalization ability. Matches generated by each method are drawn, where the red lines indicate epipolar error (pose) or projection error (homography) beyond $5\times 10^{-4}$ or $3$ pixels. Details are recorded in the top-left of each image pair, including the geometric errors created by default RANSAC estimation and the (\# correct match / \# match).}
    \label{fig2:vis-results}    \vspace{-5pt}
\end{figure*}

Existing studies focus more on RGB-only image matching due to the accessible training sets~\cite{dai2017scannet,li2018megadepth}, which have given birth to many advanced matching architectures~\cite{sarlin2020superglue,sun2021loftr,wang2024efficient,edstedt2023dkm,edstedt2024roma}. By contrast, cross-modal matching datasets are weak in scale and scene coverage, as concluded in~\cref{tab:DatasetsSta}.
The main reasons are as follows: i) It is laborious to capture a large number of multimodal images of the same target/scene, and also hard to ensure rich scene coverage. Therefore, existing datasets are often captured from driving or fixed camera views~\cite{jia2021llvip,tuzcuouglu2024xoftr}, and the number of modality types is merely two or three for each dataset.
ii) Creating precise dense labels is also expensive. Researchers often \emph{manually} label matched landmarks~\cite{jia2021llvip,liu2022target,jiang2021review}, or use camera calibration\footnote{Making two cameras as close as possible, thus considering they share a common camera pose.} to produce \emph{approximate} poses~\cite{tuzcuouglu2024xoftr,yang2024depth}.
These limited datasets can not support the training of a general matching method well due to the imbalances among them, which make the models easily dominated by simple datasets. In addition, to enlarge the data scale, researchers often generate pseudo transformations from aligned image pairs~\cite{zhangsparse,deng2024crosshomo}. However, this approach is still limited by the original data, where the stimulated deformations are not consistent with practical viewpoint changes. 
That is why existing works show poor generalization.

In this paper, we try to develop a unified matching framework for multiple cross-modal cases, by filling the data gap with an effective data engine. This engine helps us to freely scale up cheap RGB images to a large multimodal dataset with rich scenarios and accurate labels. The introduced dataset can well support the training of any matching pipelines, and largely enhance the cross-modal performance and zero-shot ability.
\textit{Our contributions are as follows}:
\begin{itemize}
  \item We are the first to develop a unified matching framework MINIMA for any cross-view and cross-modality image pairs, and achieve amazing performance enhancement.
  \item We introduce a simple yet effective data engine to freely build a high-quality multimodal dataset for image matching. Based on this, we construct MD-syn, a comprehensive dataset with large scene coverages and precise labeling, which fills the data gap for the matching community.
  \item We conduct extensive experiments on in-domain and zero-shot matching tasks including $19$ cross-modal cases, which demonstrate the high quality of our MD-syn and the promising generalization of our MINIMA.
\end{itemize}

\begin{table}[t]
 \centering
 \footnotesize
 \caption{\textbf{Overview of Representative Datasets.} It contains RGB-only and multimodal matching datasets, and our proposed MD-syn. The number (\#) of Pairs, Scene (Type: Indoor or Outdoor), Modality type, and the forms of Match Label are summarized.}
 \vspace{-5pt}
 \setlength{\tabcolsep}{1mm}{
\begin{tabular}{ccccc}
   \toprule
\textbf{Dataset} & \textbf{\#Pairs}  &  \textbf{\# Scene (Type)} & \textbf{\#Modality} & \textbf{Match Label} \\
    \midrule
    \multicolumn{5}{c}{\textit{RGB Matching}}\\
   \midrule
   \scriptsize{MegaDepth}~\cite{li2018megadepth} & 40M & 196 (Out.) & 1 & Depth, Pose \\
     ScanNet~\cite{dai2017scannet} & 230M & 1513 (In.) & 1 & Depth, Pose \\
    \midrule
    \multicolumn{5}{c}{\textit{Multimodal Matching}}\\
    \midrule
    \scriptsize{METU-VisTIR}~\cite{tuzcuouglu2024xoftr} & 2.5K  & 6 (Out.) & 2 &  Pose  \\
   M3FD~\cite{liu2022target}& 4.2K  & 15 (Out.)  & 2 &  Pre-aligned \\
    LLVIP~\cite{jia2021llvip} & 15K & 26 (Out.) & 2 & Pre-aligned  \\
    DIODE~\cite{yang2024depth}  & 25K & 20 (In. \& Out.) & 3  & Pre-aligned \\

    \scriptsize{NYU-Depth V2}\cite{silberman2012indoor} & 408k & 464 (In.) & 2 & Pre-aligned \\
     DVS128\cite{amir2017low} & 1.3k & 122 (In.) & 2 & Pre-aligned \\
    \rowcolor{gray!20} MD-syn (ours)  &  480M  & 196 (Out.) & 7 & Depth, Pose \\
   \bottomrule
\end{tabular}}
\label{tab:DatasetsSta}  \vspace{-5pt}
\end{table}

\section{Related Work}
\subsection{RGB-Only Image Matching}
Image matching is a fundamental problem in computer vision, which has aroused numerous matching methods over the past decades. Conventional pipelines start with handcrafted designs of keypoints detection, description then matching, which are recently updated with deep learning~\cite{detone2018superpoint,sarlin2020superglue,lindenberger2023lightglue,ma2021image}. These detector-based methods can establish keypoint matches with high efficiency but often struggle in textureless regions.
Recently, detector-free methods~\cite{sun2021loftr,wang2024efficient} have been introduced to produce semi-dense or dense~\cite{edstedt2023dkm,edstedt2024roma} pixel matches, and achieve dominant performance on RGB image matching in terms of match number and downstream applications. Since these methods regard each pixel as matchable points within the coarse and fine matching stages, they commonly produce a huge computational burden.
Driven by sufficient datasets, those deep methods enjoy great success in building more accurate point matches. Supported by our data engine, those advanced matching pipelines can be easily fine-tuned to multimodal cases with large enhancements.

\subsection{Multimodal Image Matching}
Image matching for multi-modalities is more challenging, due to the domain gap between two images. It often shows variations in pixel intensity distributions, making it difficult to search for matchable cues. Existing studies still rely on handcrafted designs~\cite{hou2024pos,yao2022multi,ye2019fast,li2019rift}, focusing on extracting matchable information such as shape, gradient, or phase. However, these low-level features are not consistently effective and are time-consuming to extract. Data-driven methods exhibit powerful abilities to extract matchable features for multimodal images. They commonly utilize off-the-shelf matching pipelines~\cite{deng2022redfeat,tuzcuouglu2024xoftr,liu2024grid} as the backbone, then adapt them to the target modalities with specific designs. For example,
ReDFeat~\cite{deng2022redfeat} recoupled independent
constraints of detection and description of multimodal feature
learning with a mutual weighting strategy. It performs for three cross-modality cases, but is merely trained and tested on each dataset separately.
Recently, XoFTR~\cite{tuzcuouglu2024xoftr} utilizes a two-stage training approach for RGB-IR image matching. 
It achieves large enhancement on RGB-IR image matching, by using abundant training data and a tailored matching rule.
In this paper, we contribute to filling the data gap of the general image matching. We demonstrate that our MINIMA can outperform modality-specific approaches and show superior performance in zero-shot tasks, solely relying on high-quality synthetic training data.

\subsection{Existing Datasets}

It is necessary to analyze the data gap between RGB-only image matching and the cross-modal cases.
Specifically, multi-viewed RGB images of the same target/scene are extremely cheap and easy to collect, such as directly collecting from the internet~\cite{li2018megadepth} or capturing video frames~\cite{shen2024gim}. Open-source tools like \textit{COLMAP}~\cite{schoenberger2016sfm,schoenberger2016mvs} are widely used to generate precise matching labels, such as depths and camera poses. These good datasets give birth to advanced models for RGB image matching~\cite{sarlin2020superglue,lindenberger2023lightglue,ma2021image}.
However, capturing a large number of multimodal images of the same scene is laborsome, since some imaging devices should be gathered for shooting together. This limits the scale of available image pairs. Moreover, the matching labels cannot be directly obtained by tools, which are often labeled manually~\cite{jia2021llvip,liu2022target,jiang2021review}, or approximated from camera calibrations~\cite{tuzcuouglu2024xoftr,yang2024depth}.
We conclude representative public datasets in~\cref{tab:DatasetsSta}. It shows that these multimodal datasets exhibit significant variability from each other, and are all limited by the scale and scene coverage. This impedes us from training a unified matching model for multiple cross-modal cases.

Recently, data scale-up has shown great success in general vision tasks~\cite{yang2024depth,yang2024depthv2}. They typically enlarge the training set by generating pseudo labels from huge wild RGB images. In contrast, our challenges lie in getting numerous paired images of different modalities and rich scenarios. For such purposes, we propose a data engine to generate multiple pseudo modalities from cheap RGB image pairs. On this basis, we can generate a high-quality dataset for cross-view and cross-modality image matching, that may fill the data gap for the universal matching community and will encourage more excellent matching techniques.

\begin{figure*}
  \centering
  \includegraphics[width=0.95\linewidth]{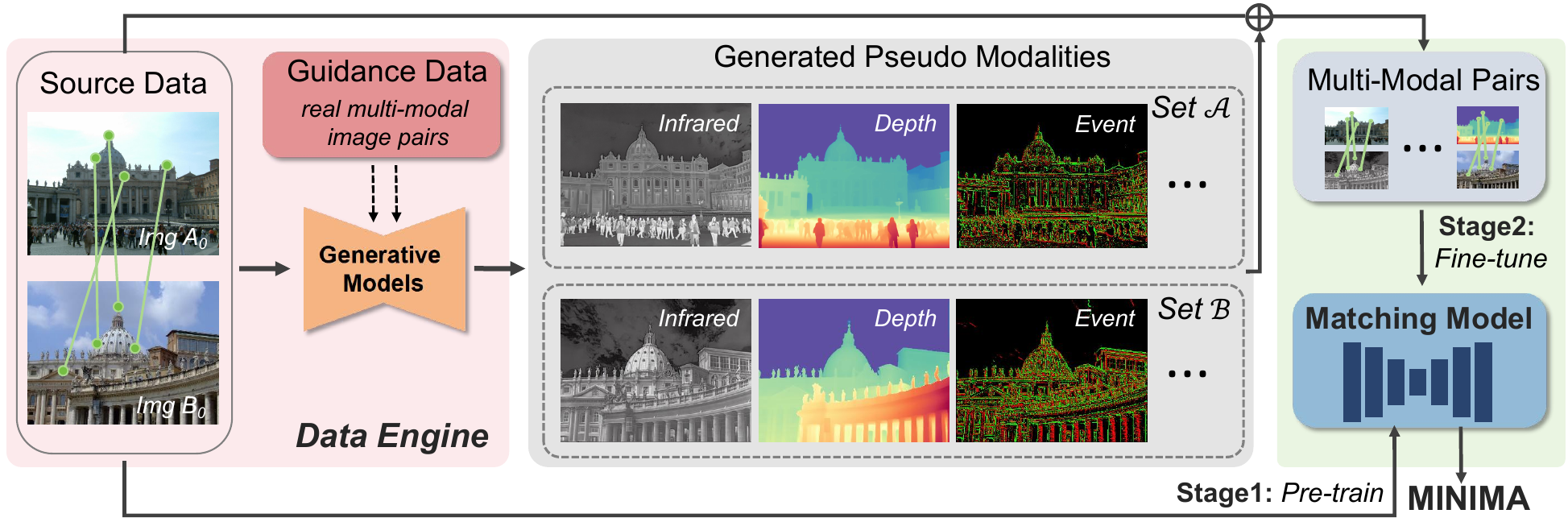}
  \caption{\textbf{Overview of the Proposed MINIMA Pipeline: Trained Once to Achieve Any Cross-modal Matching Tasks}. Wherein the \emph{Data Engine} is to generate a large multimodal matching dataset, supporting the training of matching models to obtain cross-modal ability.}\label{Fig:data-engine}  
\end{figure*}

\section{Cross-Modal Generation with Data Engine}
In this paper, we contribute to exploring a unified image matching framework for all possible image modalities by generating a large multimodal matching dataset. To achieve this, several key challenges we will face:
\begin{itemize}
 \item[-] How to obtain a large scale of image pairs with viewpoint and modality changes, and ensure the rich diversity.
 \item[-] How to freely generate dense labels of matching for those image pairs, such as depths and camera poses.
 \item[-] How to ensure the balance of different modalities in terms of scale and scene coverage.
\end{itemize}
Next, we will introduce a data engine to alleviate these concerns. It mainly benefits from the powerful ability of recent generative methods~\cite{ho2020denoising,han2024stylebooth}. The proposed engine can freely produce various pseudo modalities from real RGB image pairs, whose matching labels and scene diversity would be well inherited by the generated data.

\subsection{Advantages of Cross-Modal Generation}
The ideal strategy is to capture real images of multiple modalities in the wild. But obviously, it is impractical to arrange multiple imaging systems together. Additionally, it is more troublesome to obtain dense labels for raw image pairs, such as depth and pose information.
Another common strategy~\cite{deng2022redfeat,deng2024crosshomo,zhangsparse} involves augmenting existing aligned image pairs by randomly generating homography matrices to simulate geometric distortions. However, this is still limited by the small scale of the used dataset in diversity. The synthetic deformations are not consistent with real viewpoint changes, resulting in weak generalization of the trained model;  \emph{i.e.,} it can only work for the test set separated from the same dataset as the training set~\cite{deng2022redfeat,zhangsparse}.

To this end, we try to generate pseudo modalities to obtain a large-scale multimodal dataset, which may help to train a unified matching model for multiple cross-modal cases.
Cross-modal generation from multi-viewed RGB images has distinct advantages.
\textbf{a) Cheap:} Those RGB images are easy to collect, such as capturing from the internet~\cite{li2018megadepth} or video frames~\cite{shen2024gim}. This allows us to avoid capturing raw multimodal images in the wild.
\textbf{b) Flexible:} We can obtain any pseudo modality we want by only giving some real image pairs as guidance. With cheap RGB images, we can freely define the scale and scene of the target modality to generate. This helps us to obtain sufficient multimodal image pairs and ensure the balance of scale and scene diversity among different modalities, preventing model bias toward specific modalities.
\textbf{c) High-quality:} First, the generated images have the same resolution as RGB, breaking the limits of real sensors such as infrared or depth. Second, the matching labels of RGB images can be easily obtained by open-source tools~\cite{schoenberger2016sfm,schoenberger2016mvs}. Those accurate and dense labels can be directly inherited by the generated data.


\subsection{Scaling Up from MegaDepth}
There are many RGB image matching benchmarks, represented by MegaDepth (outdoor)~\cite{li2018megadepth} and ScanNet (indoor)~\cite{dai2017scannet}. These datasets contain millions of image pairs with depth and pose information, which are widely used and have facilitated the development of advanced matching pipelines~\cite{lindenberger2023lightglue,wang2024efficient,edstedt2023dkm,edstedt2024roma}. Here we choose MegaDepth~\cite{li2018megadepth} as the basic dataset because: i) Multimodal perception tasks are typically performed outdoors, and also, the corresponding datasets~\cite{jiang2021review} are from outdoor scenes. ii) MegaDepth demonstrates strong generalization capabilities due to its rich scene coverage and accurate labeling, making it a popular choice for training the models of existing methods~\cite{edstedt2023dkm,wang2024efficient,edstedt2024roma} to test their generalization. iii) The synthetic MegaDepth makes it convenient to fine-tune those advanced matching methods.
 Obviously, we can also generate from videos as GIM~\cite{shen2024gim}. However, GIM uses several times the scale of images but merely achieves slight gains in outdoor performance.
Considering the high computational cost of generative models, using long videos is not economical.

\subsection{Details of Our Data Engine}
We subsequently introduce how to use our data engine to generate different modalities from the public MegaDepth dataset. Here we consider the target modalities as commonly used \textit{Infrared}, \textit{Depth}, \textit{Event}, \textit{Normal}, and two \textit{Artistic Styles}. Each modality is combined with RGB to construct a cross-modal pair. Actually, we can combine any two of these modalities to form a matching pair if needed, and any other new modalities we want can also be added.

As depicted in~\cref{Fig:data-engine}, our data engine consists of three parts: \textit{Source Data, Guidance Data}, and \textit{Generative Models}. The source data is multi-viewed RGB images that we want to scale up, \emph{i.e.,} MegaDepth. The guidance data is real image pairs of our target cross-modality, mainly for fine-tuning the generative models. Here, we use publicly aligned data introduced in~\cref{tab:DatasetsSta}.
  As for generative models, we first leverage existing models to directly obtain corresponding modalities for convenience, since recent generative methods have achieved great success in image style transfer~\cite{xiang2022adversarial,han2024stylebooth} and depth or normal generation~\cite{yang2024depthv2,bae2024dsine}. As for other modalities, such as infrared, we use the guidance data to fine-tune advanced generative models~\cite{han2024stylebooth}. 
Details are as follows:

\noindent\textbf{Infrared:}
Transferring RGB to infrared is challenging due to the significant variations in their imaging systems, making existing works hard to produce satisfying results~\cite{jia2021llvip}. To this end, we turn to a diffusion-based model for help. We use StyleBooth \cite{han2024stylebooth} as the basis due to its remarkable performance in style transfer. StyleBooth was originally used to generate artistic styles controlled by an image or text description. In our study, we fine-tuned it using aligned RGB-IR image pairs from the LLVIP~\cite{jia2021llvip} and  M3FD~\cite{liu2022target} datasets. We then implement the style tuner with LoRA~\cite{hu2022lora} of rank $256$ and standardize the resolution~\footnote{To meet the resolution, we upscale the longer side of each image to $1024$ pixels, then the short side is padded with zero.} as $1024\times1024$ for both input and output. We fine-tune it on a single GPU for $210k$ steps with a fixed  $lr=1\times10^{-4}$ and batch size $2$. 

\noindent\textbf{Depth:}  We directly use DepthAnything V2~\cite{yang2024depthv2} of the official model (the large one) to generate high-quality depth images, due to the outstanding performance of monocular depth estimation and the zero-shot ability.

\noindent\textbf{Event:} The imaging principle of an event camera is simple, which has independent pixels that respond to brightness changes in their log photocurrent $L\doteq \log(I)$. Specifically, an event $e_k\doteq (\mathbf{x}_k,t_k,p_k)$ is triggered at pixel $\mathbf{x}_k\doteq (x_k, y_k)^{\top}$ and at time $t_k$ as soon as the brightness increment reaches a temporal contrast threshold $\pm C$, \emph{i.e.,}
\begin{equation}\label{eq:event1}
\triangle L(\mathbf{x}_k,t_k) \doteq L(\mathbf{x}_k,t_k) - L(\mathbf{x}_k,t_k-\triangle t_k),
\end{equation}
with $\triangle L(\mathbf{x}_k,t_k) = p_k C$,
where $C>0$, $\triangle t_k$ is the time elapsed since the last event at the same pixel, and the polarity $p_k$ is the sign of the brightness change~\cite{lichtsteiner2008128,gallego2020event}. In our study, we randomly set $C\in[0.05,0.5]$, $p_k = \pm 1$ as suggested in~\cite{gehrig2020video} to simulate varied sensors and give a random slight motion to compute the event responses.

\noindent\textbf{Normal:}
The surface normal images are directly generated with DSINE \cite{bae2024dsine}, an advanced approach that utilizes the per-pixel ray direction and recasts surface normal estimation as relative rotation estimation between pixels.

\noindent\textbf{Artistic:}
Our artistic styles include oil paint and sketch, which are implemented with open-source models, \emph{i.e.,} Paint Transformer \cite{liu2021paint} and Anime2Sketch \cite{xiang2022adversarial}, respectively. Each of them is selected for the stylistic specialization.

With the above settings, we can obtain our data engine $\{\mathcal{F}_{\theta_i}\}_{i=1}^K$ corresponding to above $K=6$ models. On this basis, and for a pair of RGB images $\{A_0, B_0\}$, we will create two image sets $\mathcal{A}=\{A_i\}_{i=1}^K$, $\mathcal{B}=\{B_i\}_{i=1}^K$ of $K$ modality types.
Since our source data MegaDepth~\cite{li2018megadepth} contains $40M$ image pairs for image matching, we will create over $480M$ cross-modal image pairs in total, with $\{A_0,B_i\}_{i=1}^K$ or $\{A_i,B_0\}_{i=1}^K$. \textit{We term the new dataset as MD-syn}. Notably, we can also create any modality pair, such as \emph{Infrared-Event}, if needed. The training and testing sets are split similarly to the original MegaDepth.

\section{Modality Invariant Image Matching Model}
After constructing MD-syn, the training of our \emph{Modality Invariant Image MAtching} (MINIMA) is easy and clear. As shown in~\cref{Fig:data-engine}, it consists of the following two stages:
\begin{itemize}
  \item \textbf{Stage $1$}: Pre-train advanced matching models on multi-view RGB data until they are converged.
  \item \textbf{Stage $2$}: Fine-tune on randomly selected cross-modal image pairs with a small learning rate.
\end{itemize}

We adopt a pre-training and then fine-tuning strategy for the following reasons. First, training from scratch on MD-syn is challenging due to the high variance among different modalities. This requires extensive iterations for convergence. By contrast, training on the RGB dataset is easy. The pre-trained models can provide good matching priors for the multimodal matching task, making it converge rapidly (verified in our supplementary). 
In addition, the training on the RGB dataset is well studied~\cite{lindenberger2023lightglue,wang2024efficient,edstedt2024roma}, whose officially trained models can directly support our fine-tuning.

Since MegaDepth has given birth to numerous matching methods with the taxonomy of sparse, semi-dense, and dense matching, we use three representative models from them as our basic models, \emph{i.e.,} LightGlue (LG)~\cite{lindenberger2023lightglue}, LoFTR~\cite{sun2021loftr}, and RoMa~\cite{edstedt2024roma}. We will fine-tune them and release our three models, termed as MINIMA$_\mathrm{{LG}}$, MINIMA$_\mathrm{{LoFTR}}$, and MINIMA$_\mathrm{{RoMa}}$. Those models will be evaluated with in-domain and zero-shot matching on synthetic and real cross-modal datasets.

\begin{table*}[htbp]
\centering 
\caption{\textbf{Full Results on Our Synthetic Dataset.} The AUC of the pose error in percentage is reported. The best and second of each category are masked as \textbf{Bold} and \underline{Underline}, respectively.}
\label{tab:synthetic_pose} \vspace{-3pt}
\setlength{\tabcolsep}{.43mm}
\footnotesize
\begin{tabular}{l l c c c c c c c c c c c c c c c c c c c c c}  %
\toprule
\multirow{2.3}{*}{\textbf{Category}} & \multirow{2.3}{*}{\textbf{Method}} & \multicolumn{3}{c}{\textbf{RGB-IR}}  & \multicolumn{3}{c}{\textbf{RGB-Depth}} & \multicolumn{3}{c}{\textbf{RGB-Normal}} & \multicolumn{3}{c}{\textbf{RGB-Event}} & \multicolumn{3}{c}{\textbf{RGB-Sketch}} & \multicolumn{3}{c}{\textbf{RGB-Paint}} \\
\cmidrule(lr){3-5}
\cmidrule(lr){6-8}
\cmidrule(lr){9-11}
\cmidrule(lr){12-14}
\cmidrule(lr){15-17}
\cmidrule(lr){18-20}
& & \textbf{@$5^\circ$} & \textbf{@$10^\circ$} & \textbf{@$20^\circ$}& \textbf{@$5^\circ$} & \textbf{@$10^\circ$} & \textbf{@$20^\circ$} & \textbf{@$5^\circ$} & \textbf{@$10^\circ$} & \textbf{@$20^\circ$} & \textbf{@$5^\circ$} & \textbf{@$10^\circ$} & \textbf{@$20^\circ$} & \textbf{@$5^\circ$} & \textbf{@$10^\circ$} & \textbf{@$20^\circ$} & \textbf{@$5^\circ$} & \textbf{@$10^\circ$} & \textbf{@$20^\circ$}\\
\midrule
\multirow{5}{*}{\textbf{Sparse}}
& SuperGlue \cite{sarlin2020superglue} & 7.49 & 17.51 & \underline{33.54} & \underline{3.06} & \underline{6.94} & \underline{13.70} & 11.53 & 24.42 & 41.85  & \underline{10.38} & \underline{23.48} & \underline{41.63} & 21.52 & 	37.99 & 	56.17 & 11.35 & 24.15 & 42.51\\
& LightGlue (LG) \cite{lindenberger2023lightglue} & 7.64 & 17.73 & 32.86 & 1.19 & 2.87 & 6.42 & \underline{12.32} & \underline{24.93} & \underline{41.86}   & 10.11 & 22.40 & 39.33 & 26.77 & 44.47 & 62.00 & \underline{13.93} & \underline{27.99} & \underline{46.16}\\
& ReDFeat \cite{deng2022redfeat} & 2.75 & 8.56 & 20.90 & 2.20 & 6.36 & 15.25 & 2.56 & 7.25 & 17.79   & 0.00 & 0.00 & 0.00 & 5.26 & 13.91 & 29.01 & 2.73 & 7.32 & 17.83\\
& GIM$_\mathrm{LG}$ \cite{shen2024gim} & \underline{8.40} & \underline{18.88} & 33.20 & 0.00 & 0.00 & 0.12 & 12.03 & 23.93 & 38.53  & 6.75 & 14.19 & 23.81 & \textbf{28.80} & \textbf{46.82} & \textbf{63.94} & 13.18 & 26.84 & 43.45 \\
\rowcolor{gray!20} \cellcolor{white}& MINIMA$_\mathrm{LG}$ & \textbf{14.74} & \textbf{30.24} & \textbf{49.22} & \textbf{16.19} & \textbf{32.53} & \textbf{51.76} & \textbf{20.47} & \textbf{37.33} & \textbf{56.17}  & \textbf{19.00} & \textbf{36.27} & \textbf{54.97} & \underline{27.51} & \underline{45.71} & \underline{63.77} & \textbf{16.39} & \textbf{32.85} & \textbf{51.65} \\
\midrule
\multirow{5}{*}{\textbf{Semi-Dense}}
& LoFTR \cite{sun2021loftr} & 5.44 & 12.58 & 24.28 & 0.13 & 0.44 & 1.88 & 5.72 & 12.07 & 23.14  & 4.90 & 12.43 & 26.45 & 37.81 & 54.82 & 69.52 & 5.93 & 12.22 & 22.19 \\
& XoFTR \cite{tuzcuouglu2024xoftr} &              \underline{17.85} &              \underline{32.21} &  \textbf{49.53} &              \underline{12.82} &              \underline{23.10} &              \underline{36.02} &              \underline{22.74} &              \underline{38.35} &              \underline{54.71}  &  \textbf{33.33} &  \textbf{51.61} &  \textbf{67.49} &     \textbf{44.18} &     \textbf{61.39} &     \textbf{75.07} &               3.73 &               7.54 &              14.48 \\
& ELoFTR \cite{wang2024efficient} &               6.73 &              14.59 &              27.36 &               0.25 &               0.79 &               3.32 &              11.20 &              21.67 &              36.86  &               9.25 &              20.39 &              37.56 &  \underline{43.86} &  \underline{61.09} &              \underline{74.84} &  \textbf{14.09} &  \textbf{25.11} &  \textbf{39.44} \\
& GIM$_\mathrm{LoFTR}$ \cite{shen2024gim} &               2.60 &               6.79 &              15.50 &               0.00 &               0.04 &               0.27 &               0.35 &               1.06 &               4.01  &               0.44 &               1.43 &               5.28 &              17.30 &              31.82 &              48.79 &               4.84 &              10.64 &              21.82 \\
\rowcolor{gray!20} \cellcolor{white}& MINIMA$_\mathrm{LoFTR}$ &  \textbf{18.07} &  \textbf{32.36} &              \underline{48.42} &              \textbf{14.70} &              \textbf{28.81} &              \textbf{46.23} &              \textbf{27.65} &              \textbf{44.26} &              \textbf{59.88}  &              \underline{18.14} &              \underline{32.74}  &              \underline{49.11} &              36.07 &              53.54 &              68.47 &               \underline{7.79} &              \underline{15.45} &              \underline{27.39} \\
\midrule
\multirow{4}{*}{\textbf{Dense}}
& DKM \cite{edstedt2023dkm} & 15.68 & 29.46 & 46.11 & 0.10 & 0.38 & 1.92 & 23.23 & 39.28 & 55.22   & 10.18 & 18.14 & 27.78 & 56.91 & 72.25 & 83.31 & 29.64 & 44.73 & 58.57 \\
& GIM$_\mathrm{DKM}$ \cite{shen2024gim} & 11.23 & 22.72 & 37.93 & 1.42 & 4.07 & 10.86 & 14.09 & 25.81 & 40.55  & 22.86 & 38.30 & 53.58 & 50.89 & 67.12 & 79.02 & 28.22 & 43.49 & 58.06 \\
& RoMa \cite{edstedt2024roma} & \underline{20.27} & \underline{35.99} & \underline{54.02} & \underline{10.21} & \underline{22.75} & \underline{39.43} & \underline{40.99} & \underline{59.48} & \underline{74.19}   & \underline{40.86} & \underline{58.87} & \underline{73.35} & \underline{58.49} & \underline{73.90} & \underline{84.80} & \textbf{41.30} & \textbf{58.36} & \textbf{72.70} \\
\rowcolor{gray!20} \cellcolor{white}& MINIMA$_\mathrm{RoMa}$ & \textbf{24.33} & 	\textbf{40.94} & 	\textbf{58.33}	 & 		\textbf{29.56}	 & \textbf{48.58} & 	\textbf{65.87}		 & 	\textbf{47.10}	 & \textbf{64.48}	 & \textbf{77.90}			 		 & \textbf{43.83}	 & \textbf{61.48}	 & \textbf{75.21}  & 	\textbf{59.17} & \textbf{74.30}	 & \textbf{84.86}		 & 	\underline{40.09}	 & \underline{57.21} & 	\underline{71.96}\\
\bottomrule
\end{tabular} 
\end{table*}

\begin{table}[tbp]
\centering
\caption{\textbf{Evaluation on Real RGB-IR Dataset (METU-VisTIR) \cite{tuzcuouglu2024xoftr} with Pose Estimation.}   The AUC of the pose error in percentage is reported. The average runtime is listed in the last column.}
\vspace{-3pt}
\setlength{\tabcolsep}{.47mm}
\footnotesize
\begin{tabular}{l l c c c c}
\toprule
\multirow{2.3}{*}{\textbf{Category}} & \multirow{2.3}{*}{\textbf{Method}} & \multicolumn{3}{c}{\textbf{Pose estimation AUC}} & \multirow{2.3}{*}{\textbf{Time}}\\
\cmidrule(lr){3-5}
& & \textbf{@$5^\circ$} & \textbf{@$10^\circ$} & \textbf{@$20^\circ$} & (ms) \\
\midrule
\multirow{9}{*}{\textbf{Sparse}}
    & RIFT \cite{li2019rift}$_{ (\text{TIP 19})}$  & 0.05 & 	0.27 & 	0.90 & 13k\\
    & SRIT \cite{li2023multimodal}$_{ (\text{ISPRS 23})}$ & 0.00	 & 0.08	 & 0.37 &  1.9k\\
    & LNIFT \cite{li2022lnift}$_{ (\text{TGRS 22})}$ & 0.02	 & 0.09	 & 0.43  &  1.2k\\
    & SuperGlue \cite{sarlin2020superglue}$_{ (\text{CVPR 20})}$ & \underline{4.30} & 	\underline{9.26} & 	\underline{17.21}  & 86.1\\
    & ReDFeat \cite{deng2022redfeat}$_{ (\text{TIP 23})}$ & 1.71	 & 4.57	 & 10.85 & 235.8\\
    & LightGlue (LG) \cite{lindenberger2023lightglue}$_{ (\text{ICCV 23})}$  & 2.17		& 5.37		& 11.21 & 57.7 \\
    & GIM$_\mathrm{LG}$  \cite{shen2024gim}$_{ (\text{ICLR 24})}$ &   2.43	& 5.85	& 10.58   & {42.9} \\
    & OmniGlue \cite{jiang2024omniglue}$_{ (\text{CVPR 24})}$  &  1.48	 & 4.13 & 	10.11 & 3k \\
\rowcolor{gray!20} \cellcolor{white}    & MINIMA$_\mathrm{LG}$  & \textbf{19.14} &	\textbf{37.17}	& \textbf{55.51}  & 58.6 \\
\midrule
\multirow{5}{*}{\textbf{Semi-Dense}}
    & LoFTR \cite{sun2021loftr}$_{ (\text{CVPR 21})}$ & 2.88  &	 6.94	 & 14.95 & 61.6\\
    & GIM$_\mathrm{LoFTR}$ \cite{shen2024gim}$_{ (\text{ICLR 24})}$  & 0.43 & 	1.06 & 	2.99 & 69.5\\
    & ELoFTR \cite{wang2024efficient}$_{ (\text{CVPR 24})}$ & 2.88 & 	7.88 & 	17.72 & 46.6\\
    & XoFTR \cite{tuzcuouglu2024xoftr}$_{ (\text{CVPR 24})}$ & \textbf{18.47} & 	\textbf{34.64} & 	\textbf{51.50} & 62.7 \\
   \rowcolor{gray!20} \cellcolor{white} & MINIMA$_\mathrm{LoFTR}$  &  \underline{15.61} & 	\underline{30.84} & 	\underline{47.87} & 71.6\\
\midrule
\multirow{4}{*}{\textbf{Dense}}
    & DKM \cite{edstedt2023dkm}$_{ (\text{CVPR 23})}$  & 6.76  & 	13.69	 & 22.53 & 485.3 \\
    & GIM$_\mathrm{DKM}$ \cite{shen2024gim}$_{ (\text{ICLR 24})}$  & 5.08	 & 12.30	 & 23.69 & 792.2 \\
    & RoMa \cite{edstedt2024roma}$_{ (\text{CVPR 24})}$ & \underline{25.61}  & 	\underline{48.12}  & 	\underline{68.37} & 639.1 \\
   \rowcolor{gray!20} \cellcolor{white} & MINIMA$_\mathrm{RoMa}$ &\textbf{37.45}	 & \textbf{60.70} & \textbf{78.00} & 633.3 \\
\bottomrule
\end{tabular}
\label{tab:Real_IR_pose} \vspace{-7pt}
\end{table}

\section{Experiments}

\subsection{Implementation Details}
We directly use the official models of LightGlue~\cite{lindenberger2023lightglue}, LoFTR~\cite{sun2021loftr} and RoMa~\cite{edstedt2024roma} as the pre-trained models, and then fine-tune them with our MD-syn.
All the models are trained on $4$ RTX $3090$ GPUs, with batch sizes being $32$, $8$, and $12$, respectively.
Based on the pre-trained models, we first individually fine-tune for each modality. The unified model is fine-tuned on randomly selected modality pairs in each iteration, which is used for the generalization ability test. Notably, we only use \textit{RGB-IR, RGB-Depth, RGB-Normal} modality pairs for training, which is sufficient to achieve satisfying performance.

\noindent\textbf{Datasets.} The used datasets include our synthetic MD-syn and $5$ multimodal datasets of real images, which contain $19$ cross-modal cases in total. In particular, 1) \textit{MD-syn} splits two scenes for test, which consist of $1500$ image pairs for each cross-modal case following the setting of the original Megadepth. It contains $6$ cross-modal cases: $3$ of them (RGB-IR/-Depth/-Normal) are used for the in-domain test, while the rest are for zero-shot evaluation. 2) \textit{METU-VisTIR}~\cite{tuzcuouglu2024xoftr} is a real RGB-IR dataset containing $2590$ real image pairs with camera poses attached. 3) \textit{DIODE}~\cite{vasiljevic2019diode} is a real RGB-Depth/Normal dataset, containing $27858$ fully aligned image pairs. 4) \textit{DSEC}~\cite{wang2023visevent} provides $60$ sequences of RGB-Event videos. Three sequences are selected as our test set, which generates 100 RGB-Event pairs for testing. We rectify the frames following the instructions of the authors to obtain aligned image pairs. To test the generalization in 5) \textit{Remote Sensing} and 6) \textit{Medical} domains, we use MMIM datasets~\cite{jiang2021review} for evaluation, where the ground truths are manually labeled matches for homography estimation. The \textit{Remote Sensing} domain consists of $7$ cross-modal cases such as Optical-SAR, Optical-Map, Optical-Depth, \emph{etc}. The \textit{Medical} domain consists of $6$ cross-modal cases such as Retina, MRI-PET, CT-SPECT, \emph{etc}.

\noindent\textbf{Evaluation Protocols.}
The used datasets exhibit different labels for matching, such as camera poses and $2$-D homography matrices. For two-view datasets, such as our MD-syn and METU-VisTIR, the recovered poses by matches are evaluated to measure the matching accuracy. We report the area under the curve (AUC) of the pose error at thresholds $\{5^{\circ},10^{\circ},20^{\circ}\}$. As for homography, similar to~\cite{sun2021loftr}, we collect the mean projection error of four corner points and report the AUC under thresholds $\{3$px,$5$px,$10$px$\}$ for evaluation. Notably, for those aligned image pairs, we impose synthetic homography matrices on one image to imitate deformations, which are finished before evaluation to maintain fairness. Then, we try to recover the homography matrix. And we uniformly resize all images with their long dimension equal to $640$. All the experiments are performed on a single RTX $3090$ GPU for accuracy and runtime tests. For all baselines, we employ the same RANSAC~\cite{fischler1981random} settings as a robust homography or pose estimator for fair comparison.

\noindent\textbf{Baselines.}
Following~\cite{wang2024efficient}, we select representative methods from the matching pipelines of sparse, semi-dense, and dense matching. 1) For sparse keypoint detection and matching methods, we choose SuperGlue~\cite{sarlin2020superglue}, LightGlue (LG)~\cite{lindenberger2023lightglue}, OmniGlue~\cite{jiang2024omniglue}, and LG-based GIM~\cite{shen2024gim} for comparison. All of them (including our MINIMA$_\mathrm{LG}$) use SuperPoint as the keypoint detector (the maximum number of extracted keypoints is set as $2048$). We also take ReDFeat~\cite{deng2022redfeat} into account as it is a deep method designed for multimodal image matching. In addition, three handcrafted multimodal matching methods, including RIFT~\cite{li2019rift}, SRIT~\cite{li2023multimodal}, LNIFT~\cite{li2022lnift}, are also used. However, we only test them (including OmniGlue) on the real RGB-IR dataset due to their poor accuracy and huge time costs. 2) Semi-dense matching methods include LoFTR~\cite{sun2021loftr}, ELoFTR~\cite{wang2024efficient}, XoFTR~\cite{tuzcuouglu2024xoftr}, and GIM$_\mathrm{LoFTR}$, where XoFTR is tailored for RGB-IR image matching. 3) As for dense matching, DKM~\cite{edstedt2023dkm}, GIM$_\mathrm{DKM}$~\cite{shen2024gim} and recent SOTA matcher RoMa~\cite{edstedt2024roma}  are used for comparison.

\begin{table}[tbp]
\centering
\footnotesize
\caption{\textbf{Evaluation on Real RGB-Depth Dataset (DIODE)~\cite{vasiljevic2019diode} with Homography Estimation.} The AUC of the projective error in percentage is reported. }
\vspace{-3pt}
\setlength{\tabcolsep}{1.27mm}
\begin{tabular}{l l c c c}
\toprule
\multirow{3}{*}{\vspace{+0.5em}\textbf{Category}} & \multirow{3}{*}{\vspace{+0.5em}\textbf{Method}} & \multicolumn{3}{c}{\textbf{Homo. estimation AUC}} \\
\cmidrule(lr){3-5}
& & {@$3$px} & {@$5$px} & {@$10$px} \\
\midrule
\multirow{5}{*}{\textbf{Sparse}}
    & SuperGlue \cite{sarlin2020superglue}$_{ (\text{CVPR 20})}$ & \underline{1.77} & 	\underline{6.83}	 & \underline{21.15}  \\
    & ReDFeat \cite{deng2022redfeat}$_{ (\text{TIP 23})}$ & 1.01 & 		4.58	 & 	16.30 \\
    & LightGlue (LG) \cite{lindenberger2023lightglue}$_{ (\text{ICCV 23})}$ & 0.79 & 	3.30 & 	11.26 \\
    & GIM$_\mathrm{LG}$  \cite{shen2024gim}$_{ (\text{ICLR 24})}$ &   0.30 & 	1.14 & 	3.65    \\
   \rowcolor{gray!20} \cellcolor{white} & MINIMA$_\mathrm{LG}$  & \textbf{8.71} & 	\textbf{26.80}	 & \textbf{55.97}   \\
\midrule
\multirow{5}{*}{\textbf{Semi-Dense}}
    & LoFTR \cite{sun2021loftr}$_{ (\text{CVPR 21})}$ & 0.97 & 	4.20	 & 15.16 \\
    & GIM$_\mathrm{LoFTR}$ \cite{shen2024gim}$_{ (\text{ICLR 24})}$ & 0.00 & 	0.25	 & 1.15\\
    & ELoFTR \cite{wang2024efficient}$_{ (\text{CVPR 24})}$ & 0.82	 & 4.09 & 	16.69  \\
    & XoFTR \cite{tuzcuouglu2024xoftr}$_{ (\text{CVPR 24})}$ & \textbf{11.03} & 	\textbf{27.24} & 	\textbf{51.60}\\
   \rowcolor{gray!20} \cellcolor{white} & MINIMA$_\mathrm{LoFTR}$  &  \underline{5.35} & 	\underline{18.65}	 & \underline{44.85}\\
\midrule
\multirow{4}{*}{\textbf{Dense}}
    & DKM \cite{edstedt2023dkm}$_{ (\text{CVPR 23})}$ & 1.29 & 	4.23	 & 11.78\\
    & GIM$_\mathrm{DKM}$ \cite{shen2024gim}$_{ (\text{ICLR 24})}$ & 1.90	 & 6.34 & 	17.96 \\
    & RoMa \cite{edstedt2024roma}$_{ (\text{CVPR 24})}$ & \underline{9.21}	 & \underline{24.64}	 & \underline{49.31} \\
  \rowcolor{gray!20} \cellcolor{white}  & MINIMA$_\mathrm{RoMa}$  & \textbf{28.98} & 	\textbf{50.88} & 	\textbf{72.54}\\
\bottomrule
\end{tabular}
\label{tab:real_depth_homograpy}  
\end{table}

\begin{table*}[ht]
\centering
\caption{\textbf{Zero-shot Matching on Real Dataset with Homography Estimation.} The AUC of the corner error in percentage is reported. The best and second of each category are masked as \textbf{Bold} and \underline{Underline}, respectively.} \vspace{-3pt}
\footnotesize
\setlength{\tabcolsep}{3.1mm}
\begin{tabular}{l l c c c c c c c c c}
\toprule
\multirow{2.3}{*}{\textbf{Category}} & \multirow{2.3}{*}{\textbf{Method}} & \multicolumn{3}{c}{\textbf{Medical}}  & \multicolumn{3}{c}{\textbf{Remote Sensing}}  & \multicolumn{3}{c}{\textbf{RGB-Event}}\\
\cmidrule(lr){3-5}
\cmidrule(lr){6-8}
\cmidrule(lr){9-11}
& & {@$3$px} & {@$5$px} & {@$10$px}& {@$3$px} & {@$5$px} & {@$10$px} & {@$3$px} & {@$5$px} & {@$10$px} \\

\midrule
\multirow{5}{*}{\textbf{Sparse}}
    & SuperGlue \cite{sarlin2020superglue} & 30.72 & 	36.18 & 	44.66 & 	\underline{18.34} & 	27.47 & 	\underline{45.59}  & 0.00	 & 0.67 & 	\underline{8.00}\\
    & LightGlue (LG) \cite{lindenberger2023lightglue} & 35.47 & 	42.37 & 	49.48 & 	16.22	 & \underline{27.51} & 	44.62 & 	0.00 & 	0.67 & 	7.02 \\
    & ReDFeat \cite{deng2022redfeat}  & \textbf{38.55} & 		\textbf{44.26} & 		\underline{50.93} & 		15.99 & 		23.95 & 		43.72 & 		 	\underline{0.55} & 		0.97 & 	6.07  \\
    & GIM$_\mathrm{LG}$ \cite{shen2024gim}  &   24.32 & 	27.88	 & 33.84	 & 11.09 & 	17.44 & 	27.18 & 	\textbf{0.57}	 & \underline{1.08}	 & 5.54 \\
\rowcolor{gray!20} \cellcolor{white}    &  MINIMA$_\mathrm{LG}$ &  \underline{37.95}  & 	\underline{44.08}	 & \textbf{52.50}	 & \textbf{23.53} & 	\textbf{38.40} & 	\textbf{58.74}  & 	0.52 & 	\textbf{2.27} & \textbf{12.82}\\
\midrule
\multirow{5}{*}{\textbf{Semi-Dense}}
    & LoFTR \cite{sun2021loftr}& 38.42 & 	43.89 & 	50.13 & 	\underline{24.13}	 & 33.80 & 	50.79 & 	0.00	 & 0.00	 & 3.59\\
    & XoFTR \cite{tuzcuouglu2024xoftr} &  \textbf{39.67}	 & \textbf{45.60}	 & \underline{52.32} & 	\textbf{27.35} & 	\textbf{39.58} & 	\underline{56.63}& 	 	0.00	 & \underline{1.37}	 & \textbf{12.64}\\
    & ELoFTR \cite{wang2024efficient} & 34.57	 & 41.66	 & 49.08 & 	16.45 & 	29.65 & 	46.74 &  	\underline{0.64} & 	1.34 & 	7.78\\
    & GIM$_\mathrm{LoFTR}$ \cite{shen2024gim}  & \underline{39.51}	 & 44.40	 & 48.94 & 	17.96	 & 27.41	 & 37.29 & 0.00 & 0.55	 & 1.19  \\
 \rowcolor{gray!20} \cellcolor{white}   & MINIMA$_\mathrm{LoFTR}$ &  \textbf{39.67} & 	\underline{45.33} & 	\textbf{52.77} & 	23.32 & 	\underline{35.18} & 	\textbf{56.81}	 &	\textbf{0.81}	 &	\textbf{2.49}	 &	\underline{11.75} \\

\midrule
\multirow{4}{*}{\textbf{Dense}}
    & DKM \cite{edstedt2023dkm}   &  \underline{39.43}	  & {45.00}  & 	51.78	  & 26.44	  & 35.82	  & 51.20		  & 0.00  & 	0.00  & 	0.00\\
    & GIM$_\mathrm{DKM}$ \cite{shen2024gim}   & 37.78	 & 43.46 & 	48.87 & 	21.19 & 	30.28 & 	47.68	& 	 	0.00 & 	0.66	 & 7.04 \\
    & RoMa \cite{edstedt2024roma}  & \textbf{39.62} & 	\underline{45.13}	 & \underline{53.75}	 & \underline{29.24} & 	\underline{40.50}	 & \underline{57.84}  & \textbf{0.85}	 & \underline{1.69}	 & \underline{10.71}			  \\
  \rowcolor{gray!20} \cellcolor{white}  &  MINIMA$_\mathrm{RoMa}$  & 39.17 & 	\textbf{45.92} & 	\textbf{57.55} & \textbf{32.55} & 	\textbf{44.68} & 	\textbf{64.38}		 & \underline{0.54} & 	\textbf{3.51}	& \textbf{17.07}\\
\bottomrule
\end{tabular}
\label{tab:real Homograpy_estimation1}  \vspace{-5pt}
\end{table*}

\subsection{Evaluate on Our MD-syn}
We first test the matching methods on MD-syn, a multimodal image matching dataset synthesized by our data engine.
\cref{tab:synthetic_pose} reports the qualitative results. It shows that our MINIMA can largely enhance the cross-modal ability of the baselines. However, we achieve weak advantages for RGB-Sketch and RGB-Paint since these two artistic modalities are more similar to RGB. As the table revealed, GIM shows poor generalization for multimodal cases, since it is overfitted on RGB videos. ReDFeat performs not well on new scenes and even fails in the event case.
   As for the LoFTR series, the original LoFTR and ELoFTR are worse than SuperGlue and LG. Because edge or shape information is more crucial for multimodal image matching, it is difficult for semi-dense methods to build matches among textureless areas.
XoFTR achieves competitive results, as it is pre-trained on sufficient multi-spectral image pairs and equipped with many advanced designs.
 As for dense matching, DKM and GIM$_\mathrm{DKM}$ perform poorly on four cross-modal cases due to the huge modality gaps among them. The original RoMa exhibits good generalization, mainly because of the use of DINOv2~\cite{oquab2023dinov2} that has seen numerous types of images during pre-training. Our MINIMA still obtains significant enhancement over RoMa. 

\subsection{In Domain Image Matching}
We next conduct in-domain tests, \emph{i.e.,} training on synthetic data but testing on real data of the same modality.
 Two real datasets are used, including RGB-IR (METU-VisTIR~\cite{tuzcuouglu2024xoftr}) for pose estimation and RGB-Depth (DIODE~\cite{vasiljevic2019diode}) for homography estimation. The results are in~\cref{tab:Real_IR_pose} and~\cref{tab:real_depth_homograpy}.

For the RGB-IR test, our MINIMA$_\mathrm{LG}$ enhances sparse matching, with AUC increasing over $400\%$. Mostly, it even beats the SOTA semi-dense method XoFTR. As for semi-dense matching, XoFTR achieves the best performance. This is attributed to its pre-training on sufficient multi-spectral image pairs, the use of an effective data augmentation strategy, and specific designs incorporated in both the training and matching stages. In dense matching, our MINIMA combined with RoMa outperforms all other pipelines consistently with large margins.
The runtime of each method is also listed. The results show that sparse and semi-dense methods (except for handcrafted methods, ReDFeat, and OmniGlue) are often more efficient due to their fewer points to match. ELoFTR is faster than the sparse methods due to its efficient designs. This trend is consistent with existing works~\cite{wang2024efficient,edstedt2024roma}.

The same trends are obtained in the RGB-Depth matching as in~\cref{tab:real_depth_homograpy}. To be specific, our semi-dense method is worse than XoFTR.  That is because depth data is more challenging, and our MINIMA is based on LoFTR, a representative but old model without any fancy designs. But we largely enhance the original LoFTR from $4.2$ to $18.65$ @$5$px. The overall performance on all real cross-modal data is concluded in~\cref{fig1:overall-performance}, which reveals the promising generalization of our MINIMA.

\subsection{Zero-shot Matching for Unseen Modality}
We next extend to zero-shot matching. 1) \textit{Medical tasks} consist of $6$ modality pairs such as Retina, CT-SPECT, \emph{etc}. 2) \textit{Remote Sensing tasks} consist of $7$ cases such as Optical-SAR, Optical-Map, \emph{etc}. 3) \textit{RGB-Event} case is from DSEC~\cite{wang2023visevent}.  1) and 2) are both from MMIM~\cite{jiang2021review} datasets.

The quantitative results are outlined in~\cref{tab:real Homograpy_estimation1}. For medical scenes, almost all the methods have close accuracy since the datasets are either too easy or too difficult. But our MINIMA$_\mathrm{LG}$ still exhibits a few advantages. As for remote sensing cases, our MINIMA achieves large gains for sparse and dense matching. While the semi-dense matcher MINIMA$_\mathrm{LoFTR}$ falls behind XoFTR for the same reason. As for the RGB-Event matching, the task is extremely challenging due to the large modality gap. Despite this, our proposed MINIMA performs good capacity for matching them.
Some qualitative results are shown in~\cref{fig2:vis-results}, which demonstrate that our MINIMA can establish a high number and ratio of correct matches for real cross-modal image pairs.

\begin{table}[!t]
\centering
\caption{\textbf{Ablation Studies.} Test on Synthetic RGB-IR, Real RGB-IR, and Real RGB-Depth data, with different training settings.}
\vspace{-8pt}
\footnotesize
\setlength\tabcolsep{0.95mm}
\begin{tabular}{lccc}
\toprule
\multirow{2}{*}{Training Strategy}  & \scriptsize{\textbf{Syn RGB-IR}} & \scriptsize {\textbf{Rel RGB-IR}}& \scriptsize{\textbf{Rel RGB-D}}\\
        & \scriptsize{AUC@$10^\circ$} & \scriptsize{AUC@$10^\circ$} & \scriptsize{AUC@$10$px} \\
\midrule
\multirow{1}{*}{Basic Model: LoFTR (LT)~\cite{sun2021loftr}} & 12.58 & 6.94 & 15.16 \\
\midrule
(1) Train from scratch on syn IR  & 23.63 & 21.41 & 30.04 \\
(2) LT + real IR & 6.28 & 9.78 & 32.93 \\
(3) LT + syn IR  & 29.43 & 29.55 & 39.23 \\
(4) LT + syn Depth & 17.30 & 15.12 & 36.06 \\
\rowcolor{gray!20}(5) LT + syn IR/Depth/Normal  & \textbf{32.36} & \textbf{30.84}& \textbf{44.85} \\
\bottomrule
\end{tabular}
\label{tab:AblationStudy}
\vspace{-10pt}
\end{table}

\subsection{Ablation Studies}
\label{sec:AblaStu}
In this part, we conduct ablation studies to analyze the superiority of our MINIMA. The results on synthetic RGB-IR, real RGB-IR, and real RGB-Depth data are reported in~\cref{tab:AblationStudy}. We use LoFTR (LT) as the basic model, which serves as the pre-trained model for (2)-(5). (1) Directly train LT on synthetic RGB-IR from scratch. (2) Fine-tune LT on real RGB-IR datasets (LLVIP and M3FD). (3) Fine-tune LT only with our synthetic RGB-IR. (4) Fine-tune LT only with our synthetic RGB-Depth. (5) Fine-tune LT on mixed data of our synthetic RGB-IR, RGB-Depth, and RGB-Normal. The results of (1) and (3) reveal that training from scratch is worse than fine-tuning. (2) and (3) demonstrate the advantages of our synthetic data against real datasets. (4) reveals that only fine-tuning on synthetic RGB-Depth can well generalize to other cross-modal cases, even better than test (2) on real RGB-IR data. (5) and (3) reveal that different synthetic data can cooperate for better performance. Our full model can largely enhance the generalization ability. 

\subsection{Discussion on Possible Limitations} 
Our objective is to generate pseudo modalities to form a large multimodal dataset, which would produce two possible limitations:
\emph{i) The gap between real and pseudo modality.}
\emph{ii) The fake information during generation.}
Fortunately, these two possible limitations have little impact on our task.
First, multimodal images intrinsically vary in pixel intensity distributions~\cite{xu2023murf}. This property is well exhibited in our generated modalities (see the suppl.), which plays an important role in training a general matching model.
Existing diffusion-based methods~\cite{ho2020denoising,han2024stylebooth} can generate high-quality images of the target modality,  making the pseudo modality much closer to the real. Extensive experiments verify the high quality of our generated data. As for the generated fake information, it can well imitate the multimodal cases, \emph{e.g.} the target is visible in infrared but not in RGB, which may help to enhance the robustness of the trained model.

\section{Conclusion}

This paper presents a unified matching framework, named MINIMA, for any cross-modal cases. It is achieved by filling the data gap using an effective data engine that freely scales up cheap RGB data into a large multimodal one. The constructed MD-syn dataset contains rich scenarios and precise match labels, and supports the training of any advanced matching models, significantly improving cross-modal performance and zero-shot ability in unseen cross-modal cases.

\noindent \textbf{Acknowledgement.} This work was supported by the National Natural Science Foundation of China (Grant 62406117, U234120202, and 62225603), the China Postdoctoral Science Foundation (Grant 2023M741263 and GZC20230895), and the Postdoctor Project of Hubei Province (Grant 2024HBBHCXA014).

{
    \small
    \bibliographystyle{ieeenat_fullname}
    \bibliography{main}
}

\clearpage

\appendixpage
\begin{appendices}
\setcounter{figure}{0}
\setcounter{table}{0}
\renewcommand{\appendixname}{Appendix~\Alph{section}}
\renewcommand\thefigure{A\arabic{figure}}    
\renewcommand\thetable{A\arabic{table}}  

We first provide more details of our data engine and the proposed MINIMA. Then we conduct additional experiments, including more ablation studies,  more quantitative and qualitative matching results, and applying our MINIMA to the \textit{Visual Localization}.

\section{Details of Our Data Generation}
\subsection{Quality Verification of Modality Generation}
\label{sec:ir gen and test}
The generation models for several modalities, excluding infrared (IR), have achieved significant success in their respective domains. Therefore, we additionally evaluate the quality of our infrared generation model.  We use a diffusion-based method~\cite{han2024stylebooth} for fine-tuning due to its significant performance in style transfer. The fine-tuning process utilizes $80\%$ real RGB-IR pairs from LLVIP~\cite{jia2021llvip} and  M3FD~\cite{liu2022target}, while the rest $20\%$ is used for the test. These two datasets provide over 10,000 real RGB-IR image pairs, which are fully aligned by manual labeling.

\begin{figure}
\centering
\includegraphics[width=0.95\linewidth]{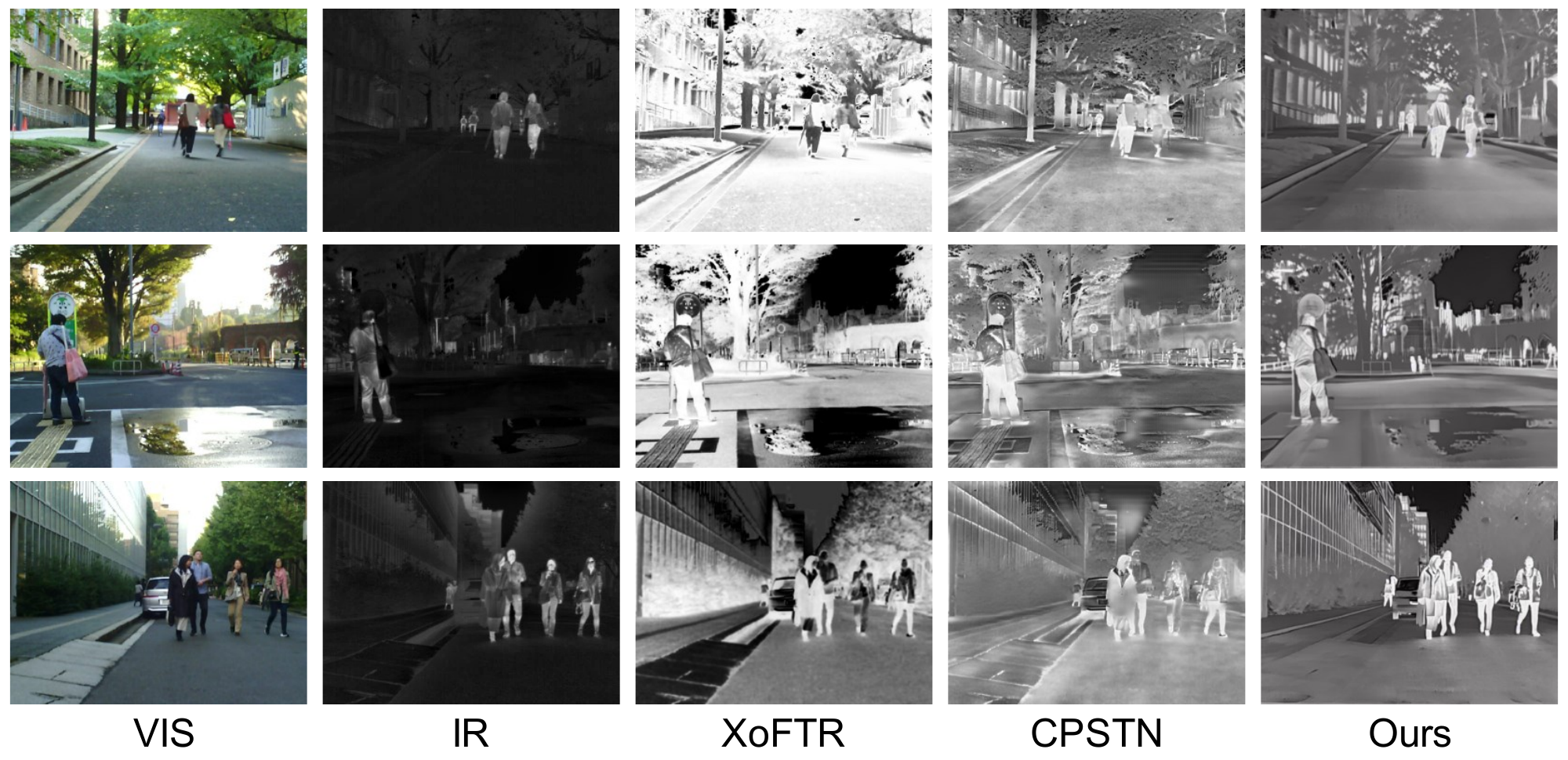}
\vspace{-7pt}
\caption{\textbf{Visualization Results of Infrared generation on MSRS}. The first two columns are real RGB and Infrared images.}
\label{Fig:MSRS_vis}
\end{figure}

\begin{figure}
\vspace{-7pt}
\centering
\includegraphics[width=0.95\linewidth]{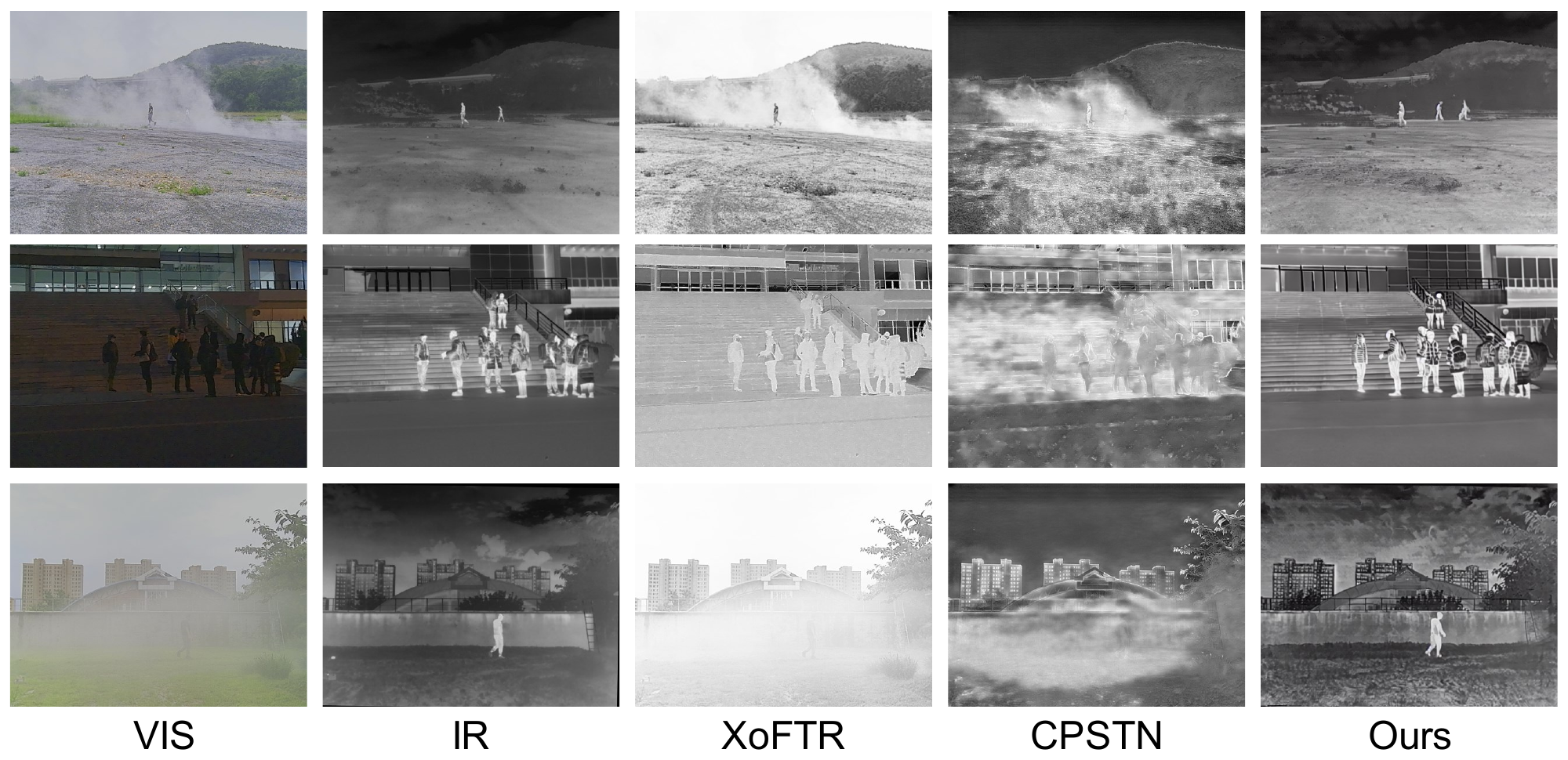}
\vspace{-7pt}
\caption{\textbf{Visualization Results of Infrared Generation on M3FD}. The first two columns are real RGB and Infrared images.}
\label{Fig:M3FD_vis}
\end{figure}

\noindent\textbf{Evaluation Protocol.} In addition to LLVIP and M3FD, we
additionally evaluate the generation performance on the MSRS dataset~\cite{Tang2022PIAFusion} by randomly selecting $120$ RGB-IR pairs. Specifically, we generate the pseudo-IR image for one RGB and then compare it with the corresponding real IR image.
For comparison, we adopt XoFTR (CVPR 24)~\cite{tuzcuouglu2024xoftr} and CPSTN (IJCAI 22)~\cite{wang2022unsupervised} as baseline methods. XoFTR used a handcrafted method to transfer RGB to IR, while CPSTN is a cycle-consistent perceptual network.
  We employ quantitative metrics including PSNR (Peak Signal-to-Noise Ratio), SSIM (Structural Similarity Index Measure) \cite{wang2004image}, LPIPS (Learned Perceptual Image Patch Similarity) \cite{zhang2018unreasonable} with AlexNet \cite{krizhevsky2012imagenet}, and PyTorch FID (Frechet Inception Distance) \cite{Seitzer2020FID}. As for FID, we compute the dimensionality of features with sizes $2048$ to serve as an evaluation indicator. The results are presented in ~\cref{tab:generation_test} with visualizations provided in ~\cref{Fig:MSRS_vis} and ~\cref{Fig:M3FD_vis}

\noindent\textbf{Results Analysis.} From both qualitative and quantitative results, we find our infrared generation achieves huge improvements. Specifically, our generated infrared images are closer to the real sensors, and the contents are clear and even better than ground truths. In addition, almost all the metrics demonstrate superiority to others by large margins. The promising performance helps a lot for our data engine to generate high-quality cross-modal image pairs.  It is also critical in training a unified matching model, making our MINIMA obtain high generalization ability.

\begin{table}[htbp]
\centering
\caption{\textbf{Quantitative Evaluation of Infrared Generation with Different Metrics}. The test datasets are LLVIP~\cite{jia2021llvip}, M3FD~\cite{liu2022target} and MSRS~\cite{Tang2022PIAFusion}. CPSTN (IJCAI 22)~\cite{wang2022unsupervised} and XoFTR (CVPR 24)~\cite{tuzcuouglu2024xoftr} are used for comparison. \textbf{Bold} indicates the best.
}
\vspace{-7pt}
\setlength{\tabcolsep}{1.8mm}
\footnotesize
\begin{tabular}{l l c c c c c}
\toprule
\multirow{1}{*}{\textbf{Data}} & \multirow{1}{*}{\textbf{Method}} & \multirow{1}{*}{\textbf{PSNR} $\uparrow$} & \multirow{1}{*}{\textbf{SSIM} $\uparrow$} & \multirow{1}{*}{\textbf{LPIPS} $\downarrow$}  & \multirow{1}{*}{\textbf{FID-2048} $\downarrow$}\\
\midrule
\multirow{3}{*}{\textbf{LLVIP}}
    & CPSTN  & 27.91 & 	0.32 & 	0.66	 	 & 303.55\\
    & XoFTR & 27.90	 & 0.29	 & 0.71		 & 204.44\\
 \rowcolor{gray!20} \cellcolor{white}   & \textbf{Ours}  &  \textbf{28.28}	 & \textbf{0.55} & 	\textbf{0.42} 	 & \textbf{145.93}\\
\midrule
\multirow{3}{*}{\textbf{M3FD}}
    & CPSTN & 27.82 & 	0.37	 & 0.56 	 & 161.71\\
    & XoFTR  & 27.86 & 	0.33	 & 0.59 	 & 125.07\\
  \rowcolor{gray!20} \cellcolor{white}  & \textbf{Ours}  &  \textbf{28.14} & 	\textbf{0.53}	 & \textbf{0.46} & \textbf{119.96}\\
\midrule
\multirow{3}{*}{\textbf{MSRS}}
    & CPSTN  & \textbf{27.95} & 	0.15	 & 0.74	 & 204.37\\
    & XoFTR  & 27.84	 & 0.16	 & 0.77	 & 167.39\\
  \rowcolor{gray!20} \cellcolor{white}  & \textbf{Ours}  &  27.87	 & \textbf{0.19} & \textbf{0.73}	 & \textbf{161.37}\\
\bottomrule
\end{tabular}
\label{tab:generation_test} \vspace{-5pt}
\end{table}

\subsection{Data Cleaning}
It is necessary to clean up the synthetic data to reduce the impacts of abnormal ones since we can not ensure the quality of the generated images.  To this end, and for each RGB image and its corresponding pseudo modalities, we use our matching model (fine-tuned on the target modality) to recover the homographies (the ground truths are the identity matrix) for them. Any image pair with the mean projection error of corner points larger than $10$ pixels is regarded as dirty data and dropped.
Finally, $ 0.91 \%$ of the matching pairs are dropped in the training set.

\begin{figure}[!tp]
  \centering
  \includegraphics[width=0.98\linewidth]{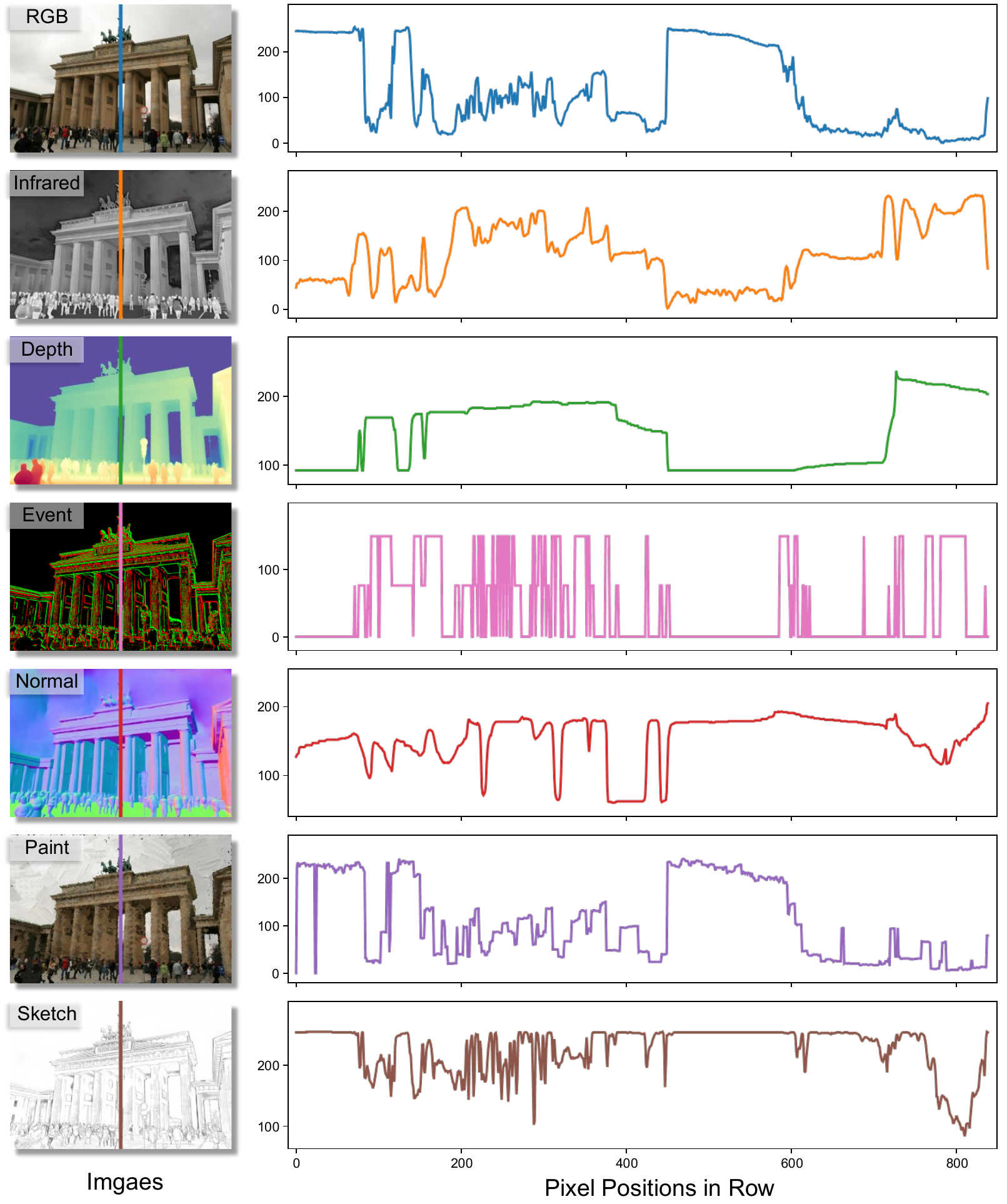} 
  \caption{ \textbf{Pixel Intensity Statistic for Generated Modalities}. The statistical differences reveal the excellent ability of our data engine to generate modality gaps. }\label{Fig:pixel-stat}  
\end{figure}

\subsection{Analysis of the Generated Modalities}
We show a group of generated modalities of our data engine in~\ref{Fig:pixel-stat}, which show high-quality visible results.  We also provide their pixel intensity statistics, which reveal the modality gap among them. These high-quality images together with their modality gaps enable existing matching models to easily obtain cross-modal ability. Not only image matching, our MINIMA can also deliver significant reference and insight for other multimodal perception tasks.

\begin{table*}[tp]
\centering \vspace{-3mm}
\caption{\textbf{Ablation Studies with Different Training Data.}  The basic models are LG, LoFTR, and RoMa. The training sets are different combinations of our generated cross-modal data. We evaluated the fine-tuned models on real cross-modal cases. For each baseline, the model trained on the original MegaDepth is reported in the first row. The average performance is shown in the last column.
} \label{tab:supp_ablation_full}
\footnotesize
\setlength{\tabcolsep}{1.15mm}
\begin{tabular}[width=\linewidth]{c|cccccc|cccccc}
\toprule
\multirow{2}{*}{\textbf{Models}} & \multicolumn{6}{c|} {\textbf{Generated Modalities}} & \textbf{Rel IR } & \textbf{Rel Depth} &\textbf{ Rel Event} & \textbf{RS} & \textbf{Medical}  & \multirow{2}{*}{\textbf{Average}}\\
 & Infrared & Depth & Normal & Event & Paint & Sketch & AUC@$10^\circ$    & AUC@$10$px          & AUC@$10$px & AUC@$10$px  & AUC@$10$px \\
\midrule
\multirow{8}{*}{MINIMA$_\mathrm{LG}$}
&&&&&&
& 5.37 &  	11.26	 &  7.02	 &  44.62	 &  49.48 &  	23.55\\
& $\checkmark$ & & & &  &
& 35.55	& 47.27	& \textbf{13.39}	& 55.12 & 	\textbf{52.73}   & 40.81\\
&  & $\checkmark$ & & & &
&  30.54	 & 51.78	 & 12.08	 & 57.73	 & 52.17  & 40.86\\
&  &  & $\checkmark$ & & &
&  32.66	 & 48.66	 & 10.44 & 	58.15	 & 52.43  & 40.47\\
&  &  &  & $\checkmark$ &  &
&  23.33	 & 38.37 & 	10.01	 & 55.72 & 	51.32 &  35.75\\
\rowcolor{gray!20} \cellcolor{white} &  $\checkmark$ & $\checkmark$ & $\checkmark$ &  & &
&  \textbf{37.17}	 & \textbf{55.97} & 12.82	 & \textbf{58.74} & 	52.50 &  \textbf{43.44}\\
\rowcolor{gray!20} \cellcolor{white} &  & &  &  $\checkmark$ & $\checkmark$ & $\checkmark$ 
  & 13.05		 & 31.09  & 
  10.89  & 
   51.52  & 
 49.67  & 

 31.24\\
\rowcolor{gray!20} \cellcolor{white} &  $\checkmark$ & $\checkmark$ & $\checkmark$ &  $\checkmark$  &  $\checkmark$ &  $\checkmark$
& 36.34	& 55.93	& 12.74	& 58.41	& 52.45	& 43.17 \\
\midrule
\multirow{9}{*}{MINIMA$_\mathrm{LoFTR}$}
&&&&&&
& 6.94	 &  15.16 &  	5.91	 &  50.79	 &  50.13	 &  25.79\\
& $\checkmark$ & & & &  &
& 29.55 &  	39.23	 &  11.12	 &  48.79 &  	51.71	 &  36.08  \\
&  & $\checkmark$ & & & &
&  15.12	 &  36.06	 &  5.32	 &  53.64 &  	52.40	 &  32.51\\
&  &  & $\checkmark$ & & &
&   23.14 &  	39.79 &  10.28	 &  54.73 &  	52.53 &  	36.09\\
&  &  &  & $\checkmark$ &  &
&  14.96 &  	32.97 &  	12.19 &  	45.77 &  	50.28	 &  31.24\\
\rowcolor{gray!20} \cellcolor{white}  &  $\checkmark$ & $\checkmark$ & $\checkmark$ & & &
&  \textbf{30.84}	 &  44.85 &  	11.38 &  	\textbf{56.81}	 &  \textbf{52.77}	 &  39.33 \\
\rowcolor{gray!20} \cellcolor{white} &  & &  &  $\checkmark$ & $\checkmark$ & $\checkmark$ 
 & 10.10 	 & 21.96 & 
11.33 & 
 	48.59  &
	49.88 & 

    35.47\\
\rowcolor{gray!20} \cellcolor{white}  & $\checkmark$ & $\checkmark$ & $\checkmark$ & $\checkmark$  & &
&  30.80	 & \textbf{48.55}	 & \textbf{12.44}	 & 56.04 & 	51.82	 & \textbf{39.93} \\
\rowcolor{gray!20} \cellcolor{white}  &  $\checkmark$ & $\checkmark$ & $\checkmark$ &  $\checkmark$  &  $\checkmark$ &  $\checkmark$
& 30.61	 & 	45.10 & 		11.83	 & 	55.33 & 		52.19 & 		39.01 \\
\midrule
\multirow{9}{*}{MINIMA$_\mathrm{RoMa}$}
&&&&&&
& 48.12	 &  49.31 &  	10.71 &  	57.84 &  	53.75	 &  43.95\\
& $\checkmark$ & & & &  &
& 57.28	 &  57.49 &  	10.49	 &  60.37	 &  57.08	 &  48.54  \\
&  & $\checkmark$ & & & &
&  60.42	 &  72.63	 &  11.00 &  	62.95 &  	56.72 &  	52.74\\
&  &  & $\checkmark$ & & &
& 60.36	 &  72.51 &  	10.89 &  	63.23 &  	55.26 &  	52.45\\
&  &  &  & $\checkmark$ &  &
&  59.11 &  	69.11 &  	11.71	 &  64.30	 &  57.75	 &  52.40 \\
\rowcolor{gray!20} \cellcolor{white}&   & $\checkmark$ & $\checkmark$ &  & &
&  58.89	 & 72.88	 & 12.36	 & 63.91 & 	55.50 & 	52.71\\
\rowcolor{gray!20} \cellcolor{white} & $\checkmark$ & $\checkmark$ & $\checkmark$ &  & &
&  60.70 &  	72.54	 &  \textbf{17.07}	 &  64.38	 &  55.09	 &  \textbf{53.96} \\
\rowcolor{gray!20} \cellcolor{white} &  & &  &  $\checkmark$ & $\checkmark$ & $\checkmark$ 
&  58.14	 & 65.91 & 
 8.79  & 
    59.66 & 
 	   57.73 & 

   50.05\\
\rowcolor{gray!20} \cellcolor{white}  &  $\checkmark$ & $\checkmark$ & $\checkmark$ &  $\checkmark$ &  &
& \textbf{61.27} &  	\textbf{73.80} &  	11.02 &  	\textbf{65.01}	 &  55.04	 &  53.23 \\
\rowcolor{gray!20} \cellcolor{white} &  $\checkmark$ & $\checkmark$ & $\checkmark$ &  $\checkmark$  &  $\checkmark$ &  $\checkmark$
& 60.43	 &  72.83 &  	12.98 &  	64.80	 &  \textbf{57.92}	 &  53.79 \\
\bottomrule
\end{tabular}\vspace{-3mm}
\end{table*}

\section{Details of MINIMA}
\label{sec:details}
The details of our fine-tuning stage are as follows. i) \textbf{LightGlue} (LG) \cite{lindenberger2023lightglue}. We use SuperPoint~\cite{detone2018superpoint} to extract $2048$ keypoints and freeze its parameters, then only fine-tune LightGlue. Because SuperPoint is verified to extract matchable features for cross-modal images~\cite{jiang2021review}. Just fine-tuning LightGlue can achieve promising performance, as demonstrated by our MINIMA$_\mathrm{LG}$. Here we directly adopted the initial learning rate, \emph{i.e.,} $1 \times 10^{-4}$, in the fine-tuning stage. In practice, we fine-tune the LG model for $50$ epochs as the authors suggested, which also shows converged performance in our study.   ii) \textbf{LoFTR}~\cite{sun2021loftr} and \textbf{RoMa}~\cite{edstedt2024roma}. We lower the learning rate to the $1/10$ of the original, with the linear scaling rule to account for batch size differences. Specifically, the initial learning rate is set as $1 \times 10^{-4}$ for LoFTR. And we set it as $1.5 \times 10^{-5}$ for the RoMa decoder, and $7.5 \times 10^{-7}$ for the RoMa encoder. Note that we maintain the default learning rate decay strategies for all methods during the fine-tuning.  For LoFTR, we fine-tune for $30$ epochs. In contrast, we fine-tune RoMa for only $4$ epochs due to its inherent capabilities, which have already achieved amazing gains. For better understanding, we also fine-tune ELoFTR~\cite{wang2024efficient} and XoFTR~\cite{tuzcuouglu2024xoftr}, denoted as MINIMA$_\mathrm{ELoFTR}$ and MINIMA$_\mathrm{XoFTR}$. And they are fine-tuned for $20$ epochs and $5$ epochs, respectively. Their learning rates are similar to our MINIMA$_\mathrm{LoFTR}$. The corresponding results are in~\cref{tab:supp_results_syn_data},~\cref{tab:supp_results_real_data} and~\cref{tab:Megadepth-rgb-1500}.

\begin{table*}[htbp]
\centering
\caption{\textbf{Semi-dense Matching Results on Our Synthetic Dataset.} The AUC of the pose error in percentage is reported. The best and second are masked as \textbf{Bold} and \underline{Underline}, respectively.} \label{tab:supp_results_syn_data}
\footnotesize
\setlength{\tabcolsep}{0.45mm}
\begin{tabular}{l l c c c c c c c c c c c c c c c c c c c c c}  %
\toprule
\multirow{2.3}{*}{\textbf{Category}} & \multirow{2.3}{*}{\textbf{Method}} & \multicolumn{3}{c}{\textbf{RGB-IR}}  & \multicolumn{3}{c}{\textbf{RGB-Depth}} & \multicolumn{3}{c}{\textbf{RGB-Normal}} & \multicolumn{3}{c}{\textbf{RGB-Event}} & \multicolumn{3}{c}{\textbf{RGB-Sketch}} & \multicolumn{3}{c}{\textbf{RGB-Paint}} \\
\cmidrule(lr){3-5}
\cmidrule(lr){6-8}
\cmidrule(lr){9-11}
\cmidrule(lr){12-14}
\cmidrule(lr){15-17}
\cmidrule(lr){18-20}
& & \textbf{@$5^\circ$} & \textbf{@$10^\circ$} & \textbf{@$20^\circ$}& \textbf{@$5^\circ$} & \textbf{@$10^\circ$} & \textbf{@$20^\circ$} & \textbf{@$5^\circ$} & \textbf{@$10^\circ$} & \textbf{@$20^\circ$} & \textbf{@$5^\circ$} & \textbf{@$10^\circ$} & \textbf{@$20^\circ$} & \textbf{@$5^\circ$} & \textbf{@$10^\circ$} & \textbf{@$20^\circ$} & \textbf{@$5^\circ$} & \textbf{@$10^\circ$} & \textbf{@$20^\circ$}\\
\midrule
\multirow{7}{*}{\textbf{Semi-Dense}}
& LoFTR \cite{sun2021loftr} & 5.44 & 12.58 & 24.28 & 0.13 & 0.44 & 1.88 & 5.72 & 12.07 & 23.14  & 4.90 & 12.43 & 26.45 & 37.81 & 54.82 & 69.52 & 5.93 & 12.22 & 22.19 \\
& XoFTR \cite{tuzcuouglu2024xoftr} &              17.85 &              32.21 &  \underline{49.53} &              12.82 &              23.10 &              36.02 &              22.74 &              38.35 &              54.71  &  \textbf{33.33} &  \textbf{51.61} &  \textbf{67.49} &     \textbf{44.18} &     \textbf{61.39} &     \textbf{75.07} &               3.73 &               7.54 &              14.48 \\
& ELoFTR \cite{wang2024efficient} &               6.73 &              14.59 &              27.36 &               0.25 &               0.79 &               3.32 &              11.20 &              21.67 &              36.86  &               9.25 &              20.39 &              37.56 &  \underline{43.86} &  \underline{61.09} &              74.84 &  \underline{14.09} &  \underline{25.11} &  \underline{39.44} \\
& GIM$_\mathrm{LoFTR}$ \cite{shen2024gim} &               2.60 &               6.79 &              15.50 &               0.00 &               0.04 &               0.27 &               0.35 &               1.06 &               4.01  &               0.44 &               1.43 &               5.28 &              17.30 &              31.82 &              48.79 &               4.84 &              10.64 &              21.82 \\
 \rowcolor{gray!20} \cellcolor{white} & MINIMA$_\mathrm{LoFTR}$ &  \underline{18.07} &  \underline{32.36} &              48.42 &              14.70 &              28.81 &              46.23 &              27.65 &              44.26 &              59.88  &              18.14 &              32.74 &              49.11 &              36.07 &              53.54 &              68.47 &               7.79 &              15.45 &              27.39 \\
\rowcolor{gray!20} \cellcolor{white} & MINIMA$_\mathrm{XoFTR}$ &     \textbf{18.97} &     \textbf{34.36} &     \textbf{51.72} &     \textbf{24.47} &     \textbf{40.90} &     \textbf{58.36} &     \textbf{30.47} &     \textbf{47.90} &     \textbf{64.64}  &     \underline{31.14} & 	\underline{49.39}	 & \underline{65.71} &              42.91 &              60.77 &  \underline{75.00} &               5.61 &              11.56 &              20.95 \\
\rowcolor{gray!20} \cellcolor{white} & MINIMA$_\mathrm{ELoFTR}$ &              13.14 &              26.36 &              43.63 &  \underline{16.59} &  \underline{32.26} &  \underline{50.37} &  \underline{29.72} &  \underline{47.47} &  \underline{63.72}  &              15.66 &              30.72 &              48.73  &              41.64 &              59.63 &              73.73 &     \textbf{15.02} &     \textbf{27.02} &     \textbf{41.62} \\
\bottomrule
\end{tabular}
\end{table*}

\begin{table*}[htbp]
\centering
\caption{\textbf{Semi-dense Matching Results on Real Dataset.} The AUC of the pose error in percentage is reported. The best and second are masked as \textbf{Bold} and \underline{Underline}, respectively.} \label{tab:supp_results_real_data}
\footnotesize
\setlength{\tabcolsep}{0.85mm}
\begin{tabular}{l l c c c c c c c c c c c c c c c c c c}  %
\toprule
\multirow{2.3}{*}{\textbf{Category}} & \multirow{2.3}{*}{\textbf{Method}} & \multicolumn{3}{c}{\textbf{Real RGB-IR}}  & \multicolumn{3}{c}{\textbf{Real RGB-Depth}} & \multicolumn{3}{c}{\textbf{Medical}} & \multicolumn{3}{c}{\textbf{Remote Sensing}} & \multicolumn{3}{c}{\textbf{Real RGB-Event}} \\
\cmidrule(lr){3-5}
\cmidrule(lr){6-8}
\cmidrule(lr){9-11}
\cmidrule(lr){12-14}
\cmidrule(lr){15-17}
& & {@$5^\circ$} & {@$10^\circ$} &{@$20^\circ$}& {@$3$px} & {@$5$px} & {@$10$px} & {@$3$px} & {@$5$px} & {@$10$px} &{@$3$px} & {@$5$px} & {@$10$px} & {@$3$px} & {@$5$px} & {@$10$px} \\
\midrule
\multirow{7}{*}{\textbf{Semi-Dense}}
& LoFTR \cite{sun2021loftr} & 2.88 & 6.94 & 14.95 & 0.97 & 4.20 & 15.16 & 38.42 & 43.89 & 50.13 & 24.13 & 33.80 & 50.79 & 0.00 & 0.00 & 3.59 \\
& XoFTR \cite{tuzcuouglu2024xoftr} & \underline{18.47} & \underline{34.64} & \underline{51.5} & \underline{11.03} & \underline{27.24} & \underline{51.60} & \textbf{39.67} & \textbf{45.60} & \underline{52.32} & \textbf{27.35} & \textbf{39.58} & \underline{56.63} & 0.00 & 1.37 & \underline{12.64} \\
& ELoFTR \cite{wang2024efficient} & 2.88 & 7.88 & 17.72 & 0.82 & 4.09 & 16.69 & 34.57 & 41.66 & 49.08 & 16.45 & 29.65 & 46.74 & \underline{0.64} & 1.34 & 7.78 \\
& GIM$_\mathrm{LoFTR}$  \cite{shen2024gim} & 0.43 & 1.06 & 2.99 & 0.00 & 0.25 & 1.15 & \underline{39.51} & 44.40 & 48.94 & 17.96 & 27.41 & 37.29 & 0.00 & 0.55 & 1.19 \\
\rowcolor{gray!20} \cellcolor{white} & MINIMA$_\mathrm{LoFTR}$ & 15.61 & 30.84 & 47.87 & 5.35 & 18.65 & 44.85 & \textbf{39.67} & \underline{45.33} & \textbf{52.77} & 23.32 & 35.18 & \textbf{56.81} & \textbf{0.81} & \textbf{2.49} & 11.75 \\
\rowcolor{gray!20} \cellcolor{white} & MINIMA$_\mathrm{XoFTR}$ & \textbf{19.38} & \textbf{35.82} & \textbf{52.94} & \textbf{11.76} & \textbf{29.48} & \textbf{55.05} & 39.33 & 44.92 & 52.09 & \underline{25.19} & \underline{37.86} & 54.36 & 0.00 & \underline{1.92} & \textbf{15.23} \\
\rowcolor{gray!20} \cellcolor{white} & MINIMA$_\mathrm{ELoFTR}$ & 12.11 & 28.07 & 47.25 & 3.96 & 16.42 & 44.03 & 39.12 & 44.61 & 52.12 & 19.70 & 33.78 & 53.83 & 0.37 & 1.04 & 9.66 \\
\bottomrule
\end{tabular}
\end{table*}

\section{Additional Experimental Results}
\label{sec:supp_exp}
\subsection{More Studies on Different Training Data}
To better understand our MINIMA, we fine-tune the basic models on different combinations of our generated cross-modal data. The obtained models are evaluated on different real scenes, and the results are reported in~\cref{tab:supp_ablation_full}. For each baseline, we first report the AUCs of the official models (without any fine-tuning) in the first row. Then we fine-tune each model on a single type of modality pair (RGB+X), which shows large enhancements compared with the basic models. Finally, we fine-tune the models on two or more modality types. The results demonstrate that different modalities can cooperate to train a better model. Using RGB-IR/Depth/Normal can achieve the best overall performance; hence, we use them as our final models. Additionally, using artistic data (Paint and Sketch) can not further enhance the performance because the artistic type has no physical property and is different from other modality types.

\subsection{More Results of Semi-dense Matching}
For a better understanding of our MINIMA, we further fine-tune ELoFTR~\cite{wang2024efficient} and XoFTR~\cite{tuzcuouglu2024xoftr} on the generated data, obtaining MINIMA$_\mathrm{ELoFTR}$ and MINIMA$_\mathrm{XoFTR}$. The corresponding results of semi-dense matching are reported in~\cref{tab:supp_results_syn_data} and~\cref{tab:supp_results_real_data}. The results reveal that with better pipelines, our MINIMA can achieve further enhancements of overall performance.

\begin{table}[htp]
\centering
\caption{\textbf{Evaluation on Original Megadepth-1500 for Pose Estimation.}   The AUC of the pose error in percentage is reported. This mainly demonstrates that our MINIMA can well preserve the RGB-only matching performance except when using LoFTR. } \label{tab:Megadepth-rgb-1500}
\vspace{-2pt}
\setlength{\tabcolsep}{1.5mm}
\footnotesize
\begin{tabular}{l l c c c}
\toprule
\multirow{2.3}{*}{\textbf{Category}} & \multirow{2.3}{*}{\textbf{Method}} & \multicolumn{3}{c}{\textbf{Pose estimation AUC}}\\
\cmidrule(lr){3-5}
& & \textbf{@$5^\circ$} & \textbf{@$10^\circ$} & \textbf{@$20^\circ$} \\
\midrule
\multirow{4}{*}{\textbf{Sparse}}
    & SuperGlue \cite{sarlin2020superglue}$_{ (\text{CVPR 20})}$ & 49.7 & 	67.1 & 	80.6\\
    & LightGlue (LG) \cite{lindenberger2023lightglue}$_{ (\text{ICCV 23})}$  & 49.9	 & 67.0	 & 80.1 \\
    & GIM$_\mathrm{LG}$  \cite{shen2024gim}$_{ (\text{ICLR 24})}$ &   41.3	 & 60.7	 & 75.9\\
  \rowcolor{gray!20} \cellcolor{white}  &   MINIMA$_\mathrm{LG}$  & 47.3	& 65.0	 & 78.6 \\
\midrule
\multirow{7}{*}{\textbf{Semi-Dense}}
    & LoFTR \cite{sun2021loftr}$_{ (\text{CVPR 21})}$ & 53.6 & 	69.9	 & 82.0\\
    & GIM$_\mathrm{LoFTR}$ \cite{shen2024gim}$_{ (\text{ICLR 24})}$  & 51.3 &	68.5	 & 81.1\\
    & ELoFTR \cite{wang2024efficient}$_{ (\text{CVPR 24})}$ & 56.4 &	72.2	 & 83.5\\
    & XoFTR \cite{tuzcuouglu2024xoftr}$_{ (\text{CVPR 24})}$ & 45.8	& 61.7	& 74.0 \\
 \rowcolor{gray!20} \cellcolor{white}    &  MINIMA$_\mathrm{LoFTR}$ & 29.9 & 	45.3 & 	59.5\\
 \rowcolor{gray!20} \cellcolor{white}    &  MINIMA$_\mathrm{ELoFTR}$  & 51.0 & 68.1	& 80.3 \\
 \rowcolor{gray!20} \cellcolor{white}    &  MINIMA$_\mathrm{XoFTR}$ & 44.5 & 60.0	& 	72.3 \\
\midrule
\multirow{4}{*}{\textbf{Dense}}
    & DKM \cite{edstedt2023dkm}$_{ (\text{CVPR 23})}$  & 60.4	 & 74.9	 & 85.1 \\
    & GIM$_\mathrm{DKM}$ \cite{shen2024gim}$_{ (\text{ICLR 24})}$  & 60.7	 & 75.5 & 	85.9 \\
    & RoMa \cite{edstedt2024roma}$_{ (\text{CVPR 24})}$ & 62.6 & 	76.7 & 	86.3 \\
 \rowcolor{gray!20} \cellcolor{white}    &  MINIMA$_\mathrm{RoMa}$ & 61.7	 & 76.5 & 	86.4 \\
\bottomrule
\end{tabular}
\end{table}

\subsection{Results on Original MegaDepth Dataset}
\label{sec: md1500 test}
In this part, we will evaluate the performance degradation on RGB-only matching tasks for those cross-modal matching methods. To this end, we test these methods back to the original MegaDepth-1500~\cite{li2018megadepth}.  We use the same settings as described in \cite{sun2021loftr,lindenberger2023lightglue}. Following previous testing, the RANSAC threshold is still set to $0.5$. For semi-dense and dense methods, the longest edge of the input images is resized to $1200$ pixels, while for sparse methods, it is resized to $1600$ pixels.  The results are summarized in~\cref{tab:Megadepth-rgb-1500}, revealing that our MINIMA can well maintain the ability of RGB-only matching, except for LoFTR.

 \begin{figure}[!htp]
\centering
\begin{subfigure}[b]{0.48\linewidth}
    \centering
    \includegraphics[width=\linewidth]{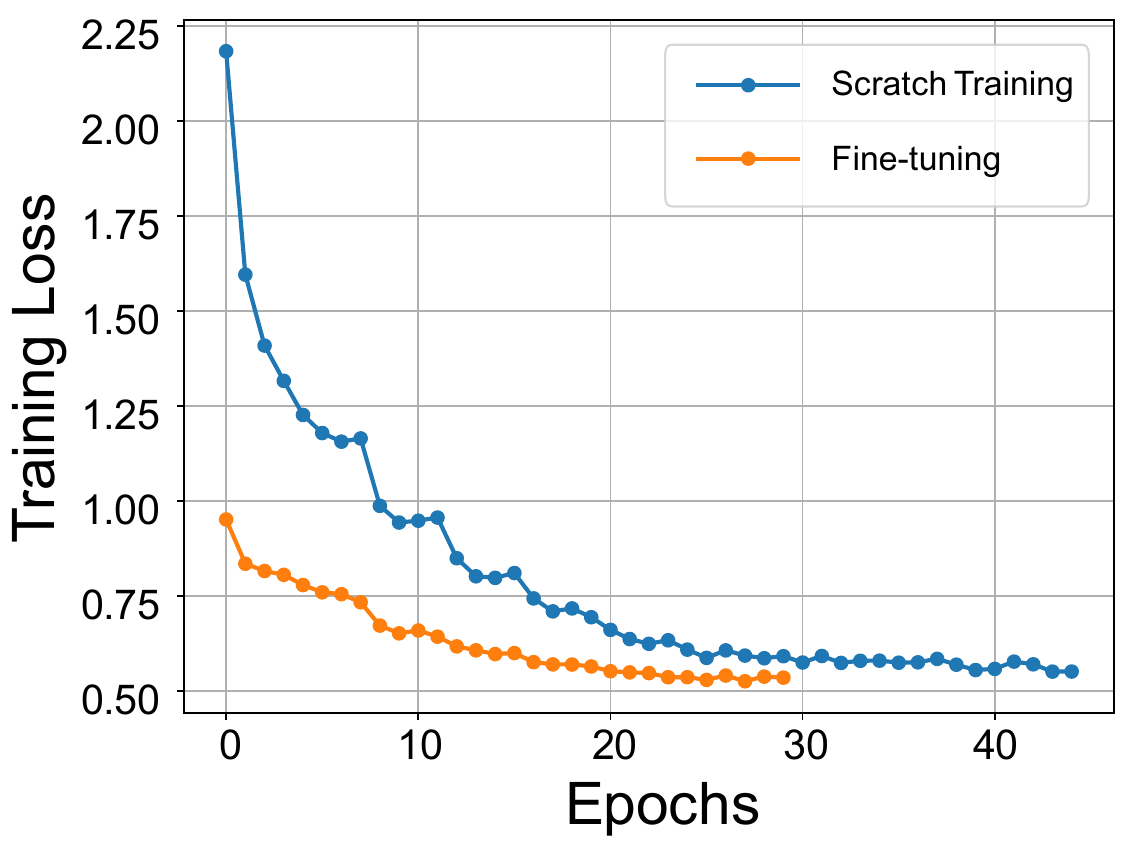}
\end{subfigure}
\begin{subfigure}[b]{0.48\linewidth}
    \centering
    \includegraphics[width=\linewidth]{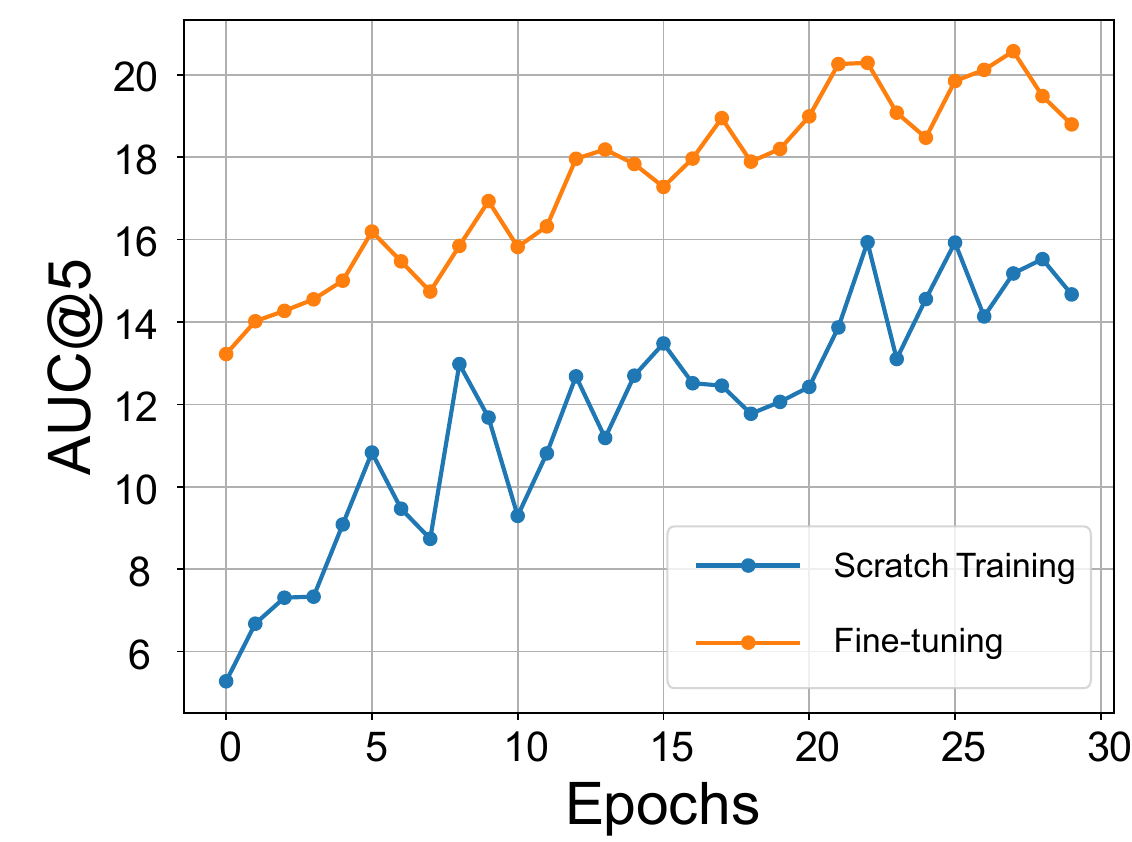}
\end{subfigure}
\vspace{-3pt}
\caption{\textbf{Training Loss and AUC@$5^\circ$ \emph{w.r.t.} Epochs, using Scratch Training and Fine-tuning}. The basic model is LoFTR. The test set is our synthetic RGB-IR of MD-syn.}
\label{Fig:Loss} \vspace{-7pt}
\end{figure}

\subsection{Scratch Training v.s. Fine-tuning}
\label{sec: scratch vs fine-tune}
 We report the loss values and AUC@$5^\circ$ performance with respect to epochs, by using scratch training and fine-tuning. The test set is our synthetic RGB-IR data generated from MegaDepth-1500~\cite{li2018megadepth}. We use LoFTR as the basic model, and the training set is our synthetic RGB-IR/Depth/Normal. Statistic results are shown in~\cref{Fig:Loss}, which reveal that the fine-tuning strategy can converge more rapidly since the pre-trained model can provide good matching priors for challenging cross-modal tasks.

\begin{table}[tp]
\centering
\caption{\textbf{Visual Localization on Aachen Day-Night V1.0}~\cite{sattler2018benchmarking}
} \label{tab:visual loc}
\vspace{-6pt}
\setlength{\tabcolsep}{1.5mm}
\footnotesize
\begin{tabular}{l c c }
\toprule
 \multirow{2.3}{*}{{\textbf{Method}}} & \textbf{Day} & \textbf{Night}\\
\cmidrule(lr){2-3}  %
& \multicolumn{2}{c}{($0.25$m,$2^\circ$) / ($0.5$m,$5^\circ$) / ({$5$m,$10^\circ$})}\\
\midrule
MNN & 86.9 / 92.0 / 95.5 &  73.5 / 79.6 / 88.8 \\
SuperGlue \cite{sarlin2020superglue}$_{ (\text{CVPR 20})}$ & 87.9 / \textbf{95.0} / 97.9  & 84.7 / \textbf{92.9} / 99.0 \\
SGMNet~\cite{chen2021learning}$_{ (\text{ICCV 21})}$  & 86.5 / 93.7 / 97.2 & 82.7 / 91.8 / 99.0 \\
LightGlue (LG) \cite{lindenberger2023lightglue}$_{ (\text{ICCV 23})}$  & 88.0 / 93.8 / 97.5	& 84.7 / 91.8 / 99.0\\
ConvMatch~\cite{zhang2023convmatch}$_{ (\text{TPAMI 23})}$ & 88.1 / 94.4 / 97.3 & 79.6 / 88.8 / 96.9 \\
\rowcolor{gray!20}  MINIMA$_\mathrm{LG}$ & \textbf{88.3} / 94.7 / \textbf{98.3}	 & \textbf{85.7} / \textbf{92.9} / \textbf{100.0}	 \\
\bottomrule
\end{tabular}
\end{table}

\subsection{Apply to Visual Localization}

\label{sec:vis loc}
Vision localization (VL) is a critical downstream task of image matching. The target is to recover the $6-$degree-of-freedom ($6-$DOF) camera pose from a query image related to a known 3D scene model. We perform it on the Aachen v1.0 dataset, which is a challenging large-scale outdoor dataset for localization with large-viewpoint and day-night illumination changes, making the localization largely rely on the robustness of matching methods. We adopt its full localization track for benchmarking.

Following~\cite{sarlin2019coarse,lindenberger2023lightglue}, we integrate different matching methods into the official HLoc pipeline~\cite{sarlin2019coarse} to achieve localization. Specifically, with COLMAP~\cite{schoenberger2016sfm,schoenberger2016mvs} toolbox, we first triangulate a 3D point cloud for all reference images with known poses and calibration, then retrieve $20$ reference images for each query image with NetVLAD~\cite{arandjelovic2016netvlad} on Aachen Day-Night v1.0. Then, we match the query image and the retrieved images with image matching methods, where the feature points are extracted up to 4096 by SuperPoint~\cite{detone2018superpoint}. Finally, the camera poses are estimated by RANSAC and a Perspective-n-Point solver. We report the pose recall at different scales of distance and angular thresholds, \emph{i.e.,}  ($0.25$m,$2^\circ$) / ($0.5$m,$5^\circ$) / ($5$m,$10^\circ$). The sparse matchers, including SuperGlue~\cite{sarlin2020superglue}, SGMNet~\cite{chen2021learning}, LightGlue (LG)~\cite{lindenberger2023lightglue}, ConvMatch~\cite{zhang2023convmatch} and our MINIMA fine-tuned with LightGlue, are used for comparison. We also report the raw results of SuperPoint directly with Mutual Nearest Neighbor (MNN) matching.

  The localization results are summarized in~\cref{tab:visual loc}, which demonstrate the good ability of our MINIMA for downstream applications. Since our MINIMA is additionally trained on high-quality multimodal image pairs, it can be more robust in complex scenarios.

\begin{figure*}[t]
    \centering
    \resizebox{\textwidth}{!}{
   \includegraphics[width=0.95\linewidth]{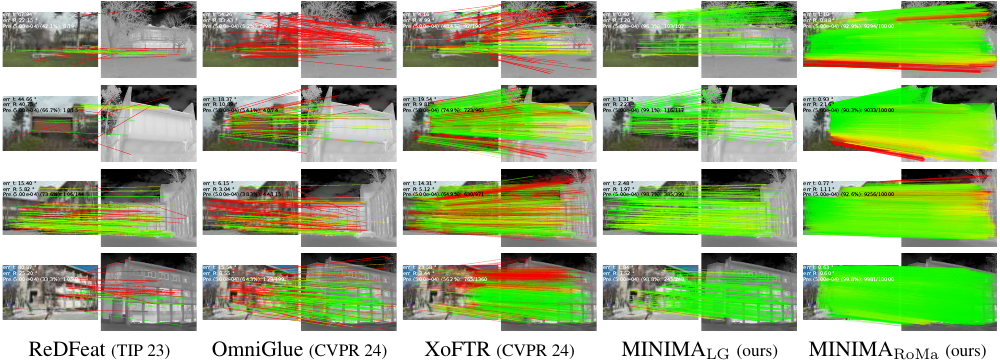}
    }
    \vspace{-15pt}
    \caption{\textbf{Qualitative Results on Real RGB-IR Image Pairs of METU-VisTIR}~\cite{tuzcuouglu2024xoftr}. The red lines indicate false matches.}
\label{Fig:IR_extra_scene}
\end{figure*}

\begin{figure*}[t]
    \centering
    \resizebox{\textwidth}{!}{
   \includegraphics[width=0.95\linewidth]{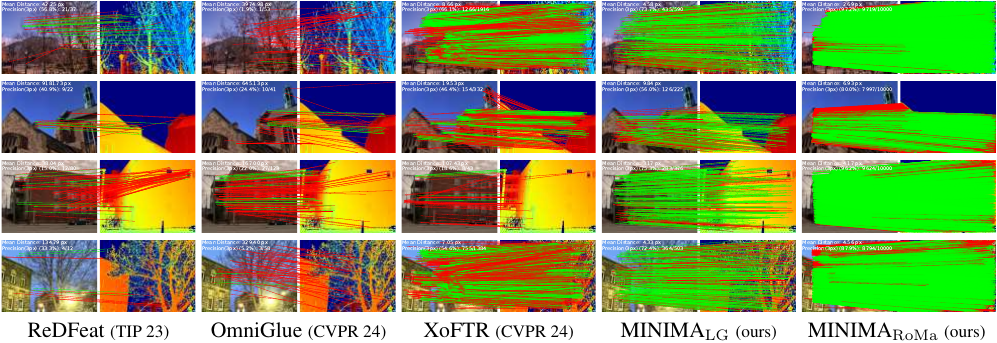}
    }
\vspace{-15pt}
\caption{\textbf{Qualitative Results on Real RGB-Depth Image Pairs of DIODE Dataset}~\cite{vasiljevic2019diode}. The red lines indicate false matches.}
\label{Fig:depth_extra_scene}
\end{figure*}

\subsection{More Visible Results on Real Datasets}
\label{sec: more vis results}
\begin{figure*}[t]
    \centering
    \resizebox{\textwidth}{!}{
   \includegraphics[width=0.95\linewidth]{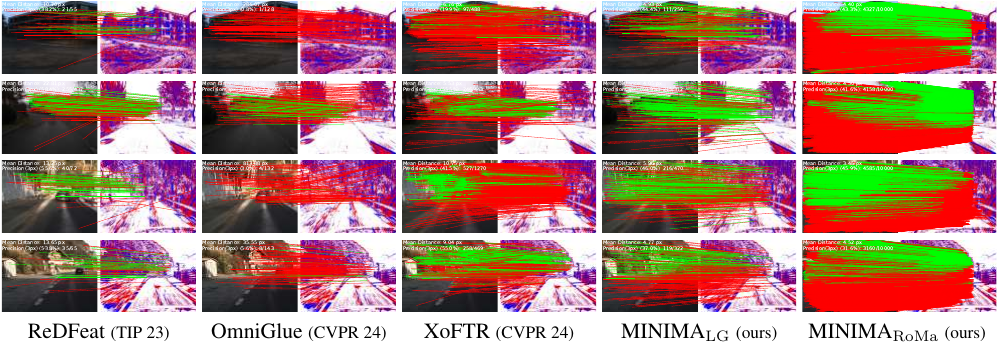}
    }
    \caption{\textbf{Qualitative Results on Real RGB-Event Image Pairs of DSEC Dataset}~\cite{wang2023visevent}. The red lines indicate false matches.}
\label{Fig:event_extra_scene}
\end{figure*}

\begin{figure*}[t]
    \centering
    \resizebox{\textwidth}{!}{
   \includegraphics[width=0.95\linewidth]{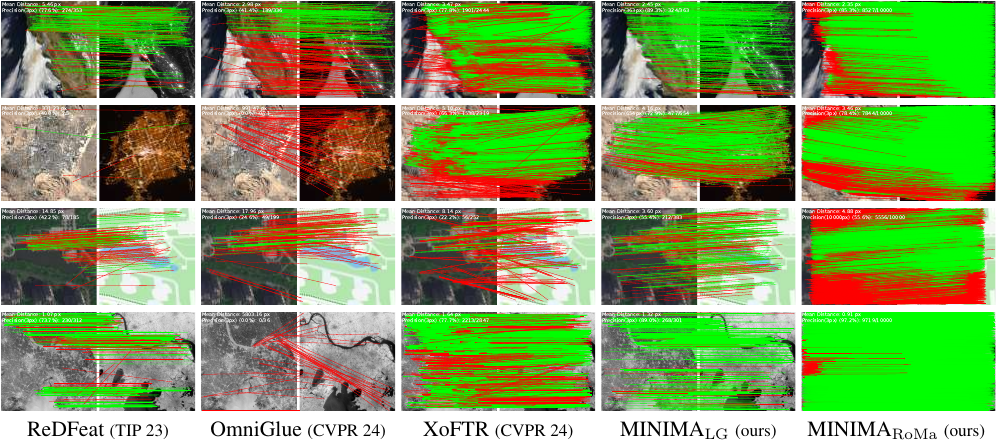}
    }
\caption{\textbf{Qualitative Results on Real Image Pairs of Cross-modal Remote Sensing Dataset}~\cite{jiang2021review}. The red lines indicate false matches.}
\label{Fig:Remote_sense_extra_scene}
\end{figure*}

We also show more qualitative results, which are selected from real RGB-IR~\cite{tuzcuouglu2024xoftr}, RGB-Depth~\cite{vasiljevic2019diode}, RGB-Event~\cite{wang2023visevent} and Remote Sensing~\cite{jiang2021review} (including Optical-SAR, optical-Map, and Day-Night) datasets. For each pair, we show the raw matching results before RANSAC. The red lines indicate false matches whose epipolar error (pose) or projection error (homography) is beyond $5\times 10^{-4}$ and $3$ pixels, respectively. Visible results are shown in~\cref{Fig:IR_extra_scene},~\cref{Fig:depth_extra_scene},~\cref{Fig:event_extra_scene} and \cref{Fig:Remote_sense_extra_scene}.
Our methods MINIMA$_\mathrm{LG}$ (sparse) and MINIMA$_\mathrm{RoMa}$ (dense) are compared with the sparse pipeline ReDFeat~\cite{deng2022redfeat} and OmniGlue~\cite{jiang2024omniglue}, and semi-dense matcher XoFTR~\cite{tuzcuouglu2024xoftr}. ReDFeat and XoFTR are cross-modal methods, and OmniGlue is known for its generalization.
The results reveal that our MINIMA can produce a high number and ratio of correct matches (green lines). 

\end{appendices}

\end{document}